\documentclass[runningheads]{llncs}

\usepackage{eccv}

\usepackage{amsfonts}
\usepackage{eccvabbrv}

\usepackage{graphicx}
\usepackage{enumitem}
\usepackage{booktabs}

\usepackage[accsupp]{axessibility}  

\usepackage{hyperref}

\usepackage{orcidlink}
\usepackage{amsmath,amssymb,amsfonts}
\usepackage{bm}

\PassOptionsToPackage{numbers}{natbib}

\usepackage[table]{xcolor}
\usepackage{algorithm}
\usepackage{algpseudocode}
\usepackage{subcaption}
\usepackage{enumitem}
\usepackage{wrapfig}
\usepackage{titletoc}
\titlecontents{ltitle}[0pt]{}{}{}{}
\titlecontents{lauthor}[0pt]{}{}{}{}
\def\Approach{Re-MeanFlow\xspace}
\newcommand{\x}{\mathbf{x}}
\newcommand{\z}{\mathbf{z}}
\definecolor{OrangeBase}{RGB}{255,128,0}
\colorlet{torange}{OrangeBase!20} 
\definecolor{toclink}{RGB}{30,120,200}

\newcommand{\bx}{\mathbf{x}}
\newcommand{\bz}{\mathbf{z}}
\newcommand{\ba}{\mathbf{a}}
\newcommand{\bb}{\mathbf{b}}

\usepackage{xcolor} 

\begin{document}

\title{Overcoming the Curvature Bottleneck in MeanFlow}

\titlerunning{\Approach{}}

\author{
Xinxi Zhang\thanks{Equal contribution}
\and
Shiwei Tan\textsuperscript{$\star$}
\and
Quang Nguyen
\and
Quan Dao
\and
Ligong Han
\and
Xiaoxiao He
\and
Tunyu Zhang
\and
Chengzhi Mao
\and
Dimitris Metaxas
\and
Vladimir Pavlovic
}

\authorrunning{X. Zhang et al.}

\institute{
Rutgers University, Piscataway NJ 08854, USA
}

\maketitle

\begin{abstract}
  MeanFlow offers a promising framework for one-step generative modeling by directly learning a mean-velocity field, bypassing expensive numerical integration. However, we find that the highly curved generative trajectories of existing models induce a noisy loss landscape, severely bottlenecking convergence and model quality. We leverage a fundamental geometric principle to overcome this: mean-velocity estimation is drastically simpler along straight paths. Building on this insight, we propose \textbf{Rectified MeanFlow}, a self-distillation approach that learns the mean-velocity field over a straightened velocity field, induced by rectified couplings from a pretrained model. To further promote linearity, we introduce a distance-based truncation heuristic that prunes residual high-curvature pairs. By smoothing the optimization landscape, our method achieves strong one-step generation performance. We improve the FID of baseline MeanFlow models from 30.9 to 8.6 under same training budget, and outperform the recent 2-rectified flow++ by 33.4\% in FID while running 26x faster. Our work suggests that the difficulty of one-step flow generation stems partially from the rugged optimization landscapes induced by curved trajectories.

  \vspace{4pt}
  \textbf{Project Page:} \url{https://xinxi-zhang.github.io/WEB_REMF/}
  
  \textbf{Code:} \url{https://github.com/Xinxi-Zhang/Re-MeanFlow}
\end{abstract}

\begin{figure}[t]
    \includegraphics[width=1.\linewidth]{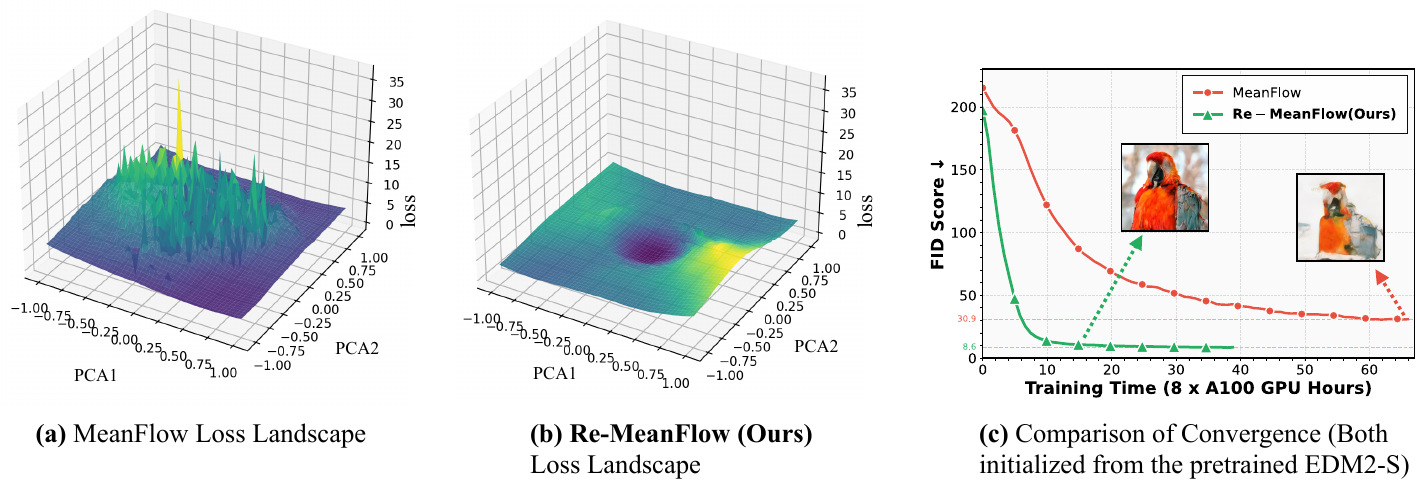}
    \caption{
        \textbf{Straightening flows smooths the loss landscape for one-step generation.}
        \textbf{(a)} MeanFlow exhibits a sharply peaked and irregular loss landscape, making optimization difficult for learning efficient one-step generators.
        \textbf{(b)} By learning the mean velocity on a rectified coupling with substantially straighter trajectories, \Approach{} (\textbf{Ours}) yields a much smoother and more regular objective.
        \textbf{(c)} This improved landscape empirically leads to faster convergence and stronger one-step generation, even when MeanFlow is trained for $2\times$ longer.
        \emph{All plots are from ImageNet $256^2$, with both MeanFlow and \Approach{} initialized from the same SiT-XL model.}
    }
    \label{fig: main}
\end{figure}

\vspace{-4pt}
\section{Introduction}
\label{sec:intro}
Flow models \cite{lipman2022flow, liu2022flow} and diffusion models \cite{song2019generative, sohl2015deep} have become a central paradigm in generative modeling, enabling a wide range of applications across various data domains \cite{ho2022video, rombach2022high, zhang2025soda,he2023dmcvr,he2024dice,xu2025can}. Compared with earlier paradigms such as GANs \cite{goodfellow2020generative, karras2019style} and Normalizing Flows \cite{rezende2015variational, zhai2024normalizing}, these models offer stable training and superior fidelity, but at the cost of expensive sampling. 
\textbf{The root cause of this inefficiency is the curvature of the generative trajectories} induced by the mismatch between the prior and data distributions. In practice, the resulting velocity field can bend sharply, making it difficult to approximate accurately with only a few discretization steps. 

This drawback has motivated works \cite{liu2022flow, tong2023improving, onken2021ot,liu2022rectified} to straighten the underlying trajectories via optimal transport; however, existing methods do not produce sufficiently linear paths, which in turn prevents reliable one-step sampling based on the instantaneous velocity.
Another line of work \cite{song2023consistency, kim2023consistency, frans2024one, geng2025mean, zhou2025inductive} bypasses ODE integration by directly learning time-indexed flow-map predictions. Among these approaches, MeanFlow \cite{geng2025mean} is particularly appealing: it models the time-dependent \emph{mean}-velocity field, enabling single-step generation without requiring straight trajectories, and achieves strong empirical performance. Despite this advantage, training MeanFlow remains costly and often converges slowly, even when initialized from a pretrained flow model.

In this work, we identify a key bottleneck in mean-velocity learning: the curvature of the underlying generative trajectories. Motivated by this insight, we introduce  \textbf{Re}ctified \textbf{MeanFlow} (\textbf{\Approach}), a lightweight self-distillation approach that trains MeanFlow on \textit{rectified} couplings, yielding markedly straighter trajectories and a simpler mean-velocity learning problem. Fig.~\ref{fig: main} highlights this contrast: optimizing MeanFlow on the original (\textit{unrectified}) trajectories produces a highly rugged landscape with sharp spikes, while \Approach yields a noticeably smoother and better-conditioned surface. \textbf{This improved conditioning leads to substantially faster convergence and superior one-step generation, even compared against MeanFlow trained with $2\times$ more compute. }
Importantly, \textbf{\Approach{} is data-free}: it requires only a pretrained flow model and samples from the prior to generate rectified couplings, without access to the original training dataset.

Additionally, we employ a simple distance-based \emph{truncation} heuristic to further reduce curvature during training. Motivated by the empirical correlation between trajectory curvature and endpoint distance, we discard the top $10\%$ of couplings ranked by their $\ell_2$ distance between endpoints. This filter removes residual high-curvature trajectories and consistently improves both training stability and sample quality in our experiments.

Extensive experiments on ImageNet at $64^2$, $256^2$, and $512^2$ show that \Approach{} \textbf{consistently outperforms} state-of-the-art distillation methods and strong train-from-scratch baselines in \emph{both} generation quality and training efficiency. In particular, relative to the closely related 2-rectified flow++~\cite{lee2024improving}, \Approach{} reduces FID by \textbf{33.4\%} while using only \textbf{10\%} of the compute.

Beyond empirical gains, \Approach{} suggests a more practical training paradigm for few-step generative models. Existing distillation pipelines rely heavily on high-end training GPUs, making hyperparameter tuning and repeated runs prohibitively expensive. In contrast, \Approach{} shifts most computation to an inference-driven reflow stage that can be executed on widely available consumer- or inference-grade accelerators, followed by a lightweight MeanFlow training phase. Overall, the training cost in \Approach{} accounts for only \textbf{17\%} of the total GPU hours used by AYF~\cite{sabour2025align}.

\section{Related Work}
We review diffusion and flow matching, emphasizing methods that reduce sampling cost via trajectory straightening or few-step modeling, and summarize recent advances in efficient training.

\subsection{Diffusion and Flow Matching.} Diffusion/flow-based generative models \cite{sohl2015deep, song2019generative, ho2020denoising, song2020score, lipman2022flow, albergo2022building, liu2022flow} learn to reverse a gradual noising process, where the reverse-time dynamics can be formulated as a deterministic probability-flow ODE \cite{song2020score, karras2022elucidating}. Although these approaches achieve strong performance, they typically require multi-step numerical integration for sampling \cite{song2020denoising, karras2022elucidating, lu2022dpm, zhang2022fast} due to the high curvature of generative paths.

\subsection{Straightening the Generative Trajectories.}
\label{sec: ot}
The high curvature of diffusion/flow generative trajectories has motivated work that explicitly \emph{reduces} the transport curvature between noise and data. OT-Flow~\cite{onken2021ot} regularizes continuous normalizing flows with OT-inspired objectives to encourage straighter dynamics. OT-CFM~\cite{tong2023improving} and Multisample Flow Matching~\cite{pooladian2023multisample} further reduce curvature by introducing OT-like \emph{minibatch} couplings, yielding simpler probability paths and lower-variance supervision. However, such minibatch couplings are inherently \emph{local} and do not directly enforce globally straighter trajectories. In contrast, Rectified Flow~\cite{liu2022flow} achieves \emph{global} trajectory straightening via reflow, iteratively refining the transport to reduce transport cost and producing near-linear paths that admit accurate few-step simulation. We therefore adopt rectified trajectories as a clean, low-curvature target for MeanFlow training, substantially easing one-step optimization.

\subsection{Few-step Diffusion/Flow Models.} Few-step models aim to bypass numerical integration by directly learning large time-step transitions. Consistency Models~\cite{song2023consistency, song2023improved, geng2024consistency, lu2024simplifying, yang2024consistency} train networks to produce invariant outputs across different timesteps, enabling direct few-step or even one-step sampling. Flow Map Models~\cite{boffi2024flow, frans2024one, geng2025mean} bypass ODE integration by learning the displacement or velocity map over time. Despite their strong empirical results, these few-step paradigms often face stability challenges or high training cost because they must learn mappings along inherently \emph{curved} trajectories, where supervision is noisy, and optimization is difficult. Recent work has sought to mitigate this issue by simplifying the consistency objective~\cite{lu2024simplifying} or introducing improved loss functions and normalization strategies\cite{dao2025improved}. Concurrent work CMT~\cite{hu2025cmt} improves the efficiency of few-step models by supervising them with teacher ODE trajectories. Building on these insights, this work aims to combine trajectory simplification with one-step modeling.

\subsection{Efficient Training for Diffusion/Flow Models.}
Many recent works \cite{yu2024representation, leng2025repa, yao2025reconstruction, wu2025representation} boost the training efficiency of diffusion models by leveraging external representation learners such as DINO \cite{oquab2023dinov2}. Orthogonal works \cite{zheng2023fast, sehwag2025stretching} adopt masked-transformer designs that exploit spatial redundancy by training only on a subset of tokens. ECT~\cite{geng2024consistency} enables efficient consistency models through a progressive training strategy that transitions from diffusion to consistency training.  \Approach{} contributes in this direction by allowing efficient one-step modeling on a significantly simplified flow path.

\section{Efficient Mean-Velocity Modeling on Straightened Trajectory}

In this section, we first review one-step flow-based generation via MeanFlow (Sec.~\ref{sec:MeanFlow}).
We then identify the \emph{curvature bottleneck}: the standard independent coupling induces highly curved paths that hinder mean-velocity learning (Sec.~\ref{sec:taming}).
Finally, we introduce \Approach{}, which models mean-velocity on rectified, substantially straighter trajectories, and describe our practical training pipeline, including CFG and distance-based truncation (Sec.~\ref{sec:train}).

\subsection{One-Step Flow-Based Generative Modeling via MeanFlow}
\label{sec:MeanFlow}

\begin{figure}[t]
    \includegraphics[width=1.\linewidth]{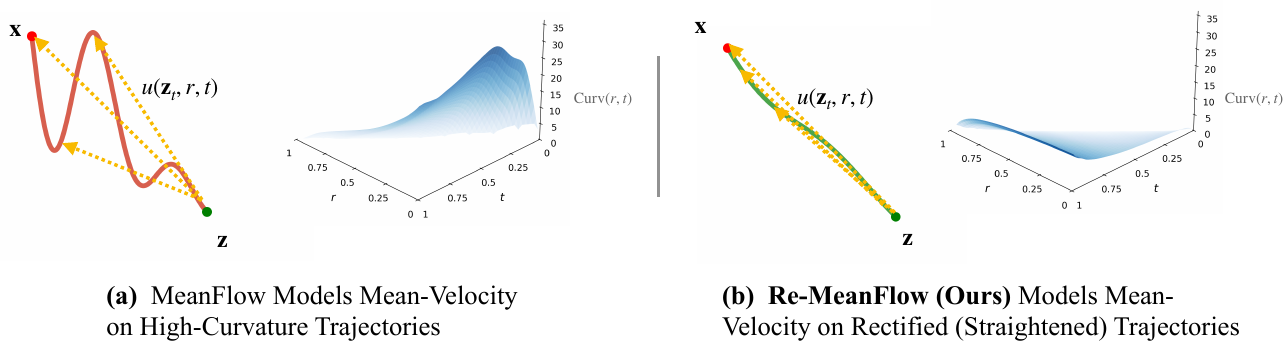}
    \caption{\textbf{Method.}
    \textbf{(a)} MeanFlow learns the mean-velocity $u(\bz_t,r,t)$ on highly curved trajectories, which makes optimization difficult. In contrast, \textbf{(b)} \textbf{\Approach{} (ours)} models $u(\bz_t,r,t)$ on a rectified coupling that induces substantially straighter generative trajectories, simplifying the underlying vector field and enabling faster and more stable training. $\mathrm{Curv}(r,t)$ denotes the curvature proxy measuring trajectory curvature between time points $r$ and $t$ (see Sec.~\ref{sec:curvature}).
    }
    \label{fig:method}
\end{figure}
Flow-based generative models transport a prior $p_{\bz}$ (taken as $\mathcal N(\mathbf 0,\mathbf I)$ in this work) to the data distribution $p_{\bx}$ by learning a time-dependent velocity field 
$v:[0,1]\times\mathbb R^d\to\mathbb R^d$ that induces a path of intermediate densities $\{p_t\}_{t\in[0,1]}$ in $\mathbb R^d$. 
Concretely, given a data-noise coupling $p_{\bx\bz}$, let $(\bx,\bz)\sim p_{\bx\bz}$ and define the linear interpolation as:
\[
\bz_t = (1-t)\bx + t\bz .
\]
Let $p_t$ denote the distribution of $\bz_t$. The coupling $p_{\bx\bz}$ together with the interpolation, therefore, specifies the probability path $\{p_t\}_{t\in[0,1]}$.

The associated velocity field $v(\bz_t,t)$ is defined as the vector field whose continuity equation
\[
\partial_t p_t + \nabla \cdot (p_t v(\cdot,t)) = 0
\]
matches this density evolution. Along the interpolation trajectory, the instantaneous velocity is given by
\[
\frac{d}{dt}\bz_t = \bz - \bx .
\]

Flow Matching (FM)~\cite{lipman2022flow,liu2022flow} models the time-dependent velocity field with a neural vector field \(v_\theta(\cdot,t)\) and trains it by regressing the trajectory velocity induced by the chosen coupling:
\begin{align}
\label{eq:fm-loss}
    &\mathcal{L}_\text{FM}(\theta) = \mathbb E_{(\mathbf x,\mathbf z)\sim p_{\mathbf x\mathbf z},\ t\sim p(t)} \left[\left\lVert\frac{\mathrm d }{\mathrm d t} \z_t- v_\theta(\mathbf z_t,t)\right\rVert_2^2\right]
\end{align}
Once $v_\theta$ is learned, a new sample $\mathbf x$ can be generated by solving the ODE for $\mathbf z \sim p_\mathbf z$ using a numerical solver:
\begin{equation}
\label{eq:fm-sampling}
    \mathbf{x}_\theta(\mathbf{z}) = \mathbf z_0 = \mathbf z - \int_0^1 v_\theta(\mathbf z_\tau, \tau) d\tau.
\end{equation}

MeanFlow~\cite{geng2025mean} instead parameterizes the \emph{mean}-velocity between two time points \(r<t\):
\begin{equation}
\label{eq:MeanFlow-def}
u(\bz_t,r,t)\triangleq \frac{1}{t-r}\int_r^t v(\bz_\tau,\tau)\,d\tau,
\end{equation}
To train a neural network $u_\theta(\mathbf z_t, r, t)$ to approximate this mean-velocity, MeanFlow derives an implicit training target that connects the mean-velocity, the instantaneous velocity $v(\mathbf z_t, t)$, and the time derivative of $u_\theta$:
\begin{equation}
    \mathcal{L}_{MF}(\theta) = \mathbb{E}_{(\mathbf x,\mathbf z)\sim p_{\mathbf x\mathbf z},\,(r,t)\sim p_{r,t}}\left\lVert u_\theta(\mathbf z_t, r, t) - \operatorname{sg}(u_{\text{tgt}})\right\rVert^2_2
    \label{eq:MeanFlow-loss}
\end{equation}
\begin{equation}
    \text{with }u_\text{tgt} = v(\mathbf z_t, t) - (t - r)\frac{d}{dt} u_\theta(\mathbf z_t, r, t).
    \label{eq:utgt}
\end{equation}
Here, $\operatorname{sg}(\cdot)$ denotes the stop-gradient operator. Once $u_\theta$ is trained, sampling reduces to a single evaluation $u_\theta(\bz,0,1)$, avoiding the numerical integration required by Flow Matching (cf.~Eq.~\ref{eq:fm-sampling}).

\subsection{Taming the Curvature Bottleneck}
\label{sec:taming}
Crucially, the underlying generative trajectories of MeanFlow are dictated by the data-noise coupling distribution $p_{\bx\bz}$. In the absence of prior knowledge about how data and noise should be paired, MeanFlow (and many flow-based methods) adopts the independent coupling distribution
\(
p^{0}_{\bx\bz}(\bx,\bz)=p_{\bx}(\bx)\,p_{\bz}(\bz),
\)
which is known to induce highly curved generative trajectories (Fig.~\ref{fig:method}a).

We argue that learning the mean-velocity field on such curved trajectories is challenging: curvature amplifies the complexity of the target mean-velocity, which induces a rugged and poorly conditioned loss landscape (visualized in Fig.~\ref{fig: main}a and Fig.~\ref{fig:loss-landscape} (top)), amplifying the effect of imperfect supervision and slowing down optimization (see Sec.~\ref{sec:loss-landscape} for further analysis). To address this bottleneck, we propose \textbf{Rectified MeanFlow} (\textbf{\Approach{}}), which models the mean-velocity field on a rectified coupling that induces substantially straighter trajectories (Fig.~\ref{fig:method}b). This rectification simplifies the underlying vector field and, correspondingly, yields a markedly smoother and more regular objective (Fig.~\ref{fig: main}b and Fig.~\ref{fig:loss-landscape} (bottom)), enabling faster convergence and improved one-step generation.

To construct this coupling that induces straighter trajectories, we follow Rectified Flow~\cite{liu2022flow} and perform a single reflow step using a pretrained flow model $v_\phi$ trained under the independent coupling. Concretely, we sample $\bz\sim p_{\bz}$ and define
\(
\bx \;=\; \bz - \int_{0}^{1} v_\phi(\bz_\tau,\tau)\, d\tau,
\)
which induces a new coupling distribution $p^{1}_{\bx\bz}(\bx,\bz)$. We then train \Approach{} by modeling the mean-velocity field on $p^{1}_{\bx\bz}$.

\subsection{Estimating the Curvature of the Generative Trajectory}
\label{sec:curvature}
To quantify the trajectory curvature over time pairs $(r,t)$ , we propose an angle-based proxy. Let $\tilde{u}(\bz_t,r,t)$ denote the \emph{mean}-velocity over the interval $[r,t]$. In practice, we approximate this quantity by integrating a learned velocity field $v_\theta$ trained under the corresponding coupling:
\begin{equation}
\tilde{u}(\bz_t,r,t)\;=\;\frac{1}{t-r}\int_{r}^{t} v_\theta(\bz_\tau,\tau)\, d\tau,
\end{equation}
and let $v_\theta(\bz_t,t)$ denote the instantaneous velocity at $(\bz_t,t)$. We then define
\begin{equation}
\label{eq:curvature-proxy}
\mathrm{Curv}(r,t)\;=\;\mathbb{E}_{\bz_t\sim p_t}\Big[\angle\!\big(\tilde{u}(\bz_t,r,t),\,v_\theta(\bz_t,t)\big)\Big],
\end{equation}
where $\angle(\ba,\bb)=\arccos\!\left(\frac{\langle \ba,\bb\rangle}{\|\ba\|\,\|\bb\|}\right)$.
Intuitively, $\mathrm{Curv}(r,t)$ measures \emph{directional disagreement} between (i) the \emph{average} transport direction over the interval $[r,t]$, given by $\tilde{u}(\bz_t,r,t)$, and (ii) the \emph{instantaneous} flow direction at time $t$, given by $v_\theta(\bz_t,t)$. 
This proxy serves as a practical diagnostic similar to the $\ell_2$-based curvature measures used in Rectified Flow~\cite{liu2022flow}; in particular, for perfectly straight trajectories, we have $\mathrm{Curv}(r,t)=0$ for all valid $(r,t)$. We visualize $\mathrm{Curv}(r,t)$ in the ImageNet-$512^2$ over the $(r,t)$-plane in Fig.~\ref{fig:method} and show that the rectified couplings used by \Approach{} exhibit substantially lower trajectory curvature.
More results at other resolutions are provided in Appendix~\ref{sec: curv_ab}.

\begin{algorithm}[t]
\caption{\Approach}
\label{alg:reMeanFlow}
\begin{algorithmic}[1]
\State \textbf{Input:} pretrained rectified flow $v_{\phi}$; prior $p_{\bz}$; ODE solver \textsc{SamplePair};
time distribution $p_{r,t}$; guidance strength $\omega$; \#pairs $N$; truncation ratio $k\%$; training iterations $T$; learning rate $\eta$.
\State \textbf{Output:} trained \Approach model $u_{\theta}$.

\State Initialize $\theta \leftarrow \textsc{InitFrom}(\phi)$ \Comment{Initialize $u_\theta$ from teacher $v_\phi$}

\Statex \textbf{Stage A: Sample rectified couplings}
\State $\mathcal P \leftarrow \{(\bx_i,\bz_i)\}_{i=1}^{N}$ where $(\bx,\bz)\sim \textsc{SamplePair}(v_\phi,p_{\bz})$
\State Compute distances $d_i \leftarrow \|\bx_i-\bz_i\|_2$ for all pairs
\State $q \leftarrow \text{Percentile}_{(100-k)}(\{d_i\}_{i=1}^{N})$
\State $\mathcal P_{\text{keep}} \leftarrow \{(\bx_i,\bz_i)\in\mathcal P : d_i \le q\}$ \Comment{distance-based truncation}

\Statex \textbf{Stage B: Train \Approach on $\mathcal P_{\text{keep}}$}
\For{$s = 1,\dots,T$}
    \State Sample $(\bx,\bz)\sim\mathcal P_{\text{keep}}$, $(r,t)\sim p_{r,t}$
    \State $\bz_t \leftarrow (1-t)\bx + t\bz$
    \State $u_{\text{tgt}} \leftarrow (\bz-\bx) - (t-r)\,\frac{d}{dt}u_{\theta}(\bz_t,r,t)$ 
    \State $\mathcal L_{\text{MF}}(\theta) \leftarrow \big\|u_{\theta}(\bz_t,r,t)-\operatorname{sg}(u_{\text{tgt}})\big\|_2^2$
    \State $\theta \leftarrow \theta - \eta \nabla_{\theta}\mathcal L_{\text{MF}}(\theta)$
\EndFor

\Statex \textbf{Stage C: CFG fine-tuning}
\State Fine-tune $u_{\theta}$ on $v^{\text{cfg}}_\omega$ (Eq.~\ref{eq:cfg}) using the loop in Stage B.

\State \textbf{Return} $u_{\theta}$
\end{algorithmic}
\end{algorithm}

\vspace{-5pt}
\subsection{Training}
\label{sec:train}

Following prior work~\cite{liu2022flow, lee2024improving}, we first construct the rectified coupling distribution $p^{1}_{\bx\bz}(\bx,\bz)$ by generating $N$ data-noise pairs with a pretrained flow model $v_\phi$, to ensure sufficient coverage of both the data and noise distributions.

\subsubsection{Classifier-free guidance.}
To enable classifier-free guidance (CFG) at inference, \Approach{} (as in MeanFlow) must be trained to predict the \emph{CFG mean-velocity} field $v_\omega^{\text{cfg}}$:
\begin{equation}
\label{eq:cfg}
    v^{\text{cfg}}_\omega(\bz_t, t \mid c) \triangleq \omega\, v(\bz_t, t \mid c) + (1 - \omega)\, v(\bz_t, t),
\end{equation}
where $\omega$ denotes the guidance strength and $c$ is the conditioning signal. In practice, we find it more stable to apply guidance via a two-stage procedure: we first train without guidance to initialize the model, and then perform a brief fine-tuning stage using the CFG objective. Overall, \Approach{} converges rapidly, typically within a single pass over the generated couplings.

\begin{figure}[t]
    \includegraphics[width=1.\linewidth]{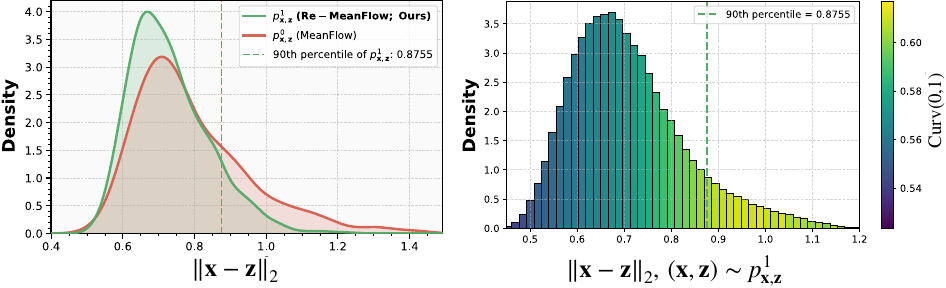}

    \caption{
\textbf{Left:} Distance statistics of the data-noise pair distance \(\|\bx-\bz\|_2\) under the independent coupling \(p^0_{\bx\bz}\) (red) and the rectified coupling \(p^1_{\bx\bz}\) (green).
\textbf{Right:} Histogram of \(\|\bx-\bz\|_2\) for rectified pairs \((\bx,\bz)\!\sim\!p^1_{\bx\bz}\), with bins colored by the curvature proxy (Sec.~\ref{sec:curvature}).
Overall, rectification reduces the average pair distance between data and noise, but a small long-distance tail remains; these \textbf{long-distance couplings tend to exhibit higher curvature}, motivating our distance-based truncation to remove many high-curvature pairs and further improve training stability and performance.
}

\label{fig:trunc1}
\end{figure}

\subsubsection{Distance-based truncation.}
\label{sec:trunc1}
Rectified Flow~\cite{liu2022flow} guarantees that rectification does not increase transport cost: for any convex cost $c:\mathbb{R}^d\!\to\!\mathbb{R}$,
\begin{equation}
\mathbb{E}_{p^{1}_{\bx\bz}}\!\left[c(\bx,\bz)\right]\;\le\;\mathbb{E}_{p^{0}_{\bx\bz}}\!\left[c(\bx,\bz)\right].
\end{equation}
We empirically verify this with $c(\bx,\bz)=\|\bx-\bz\|_2$ and plot the resulting distance distributions in Fig.~\ref{fig:trunc1} (left).
As expected, the rectified coupling $p^1_{\bx\bz}$ is \emph{left-shifted} relative to the original coupling $p^0_{\bx\bz}$, indicating smaller transport distances. 
Notably, however, $p^1_{\bx\bz}$ still exhibits a non-negligible long-tail subset of rectified pairs $(\bx,\bz)\!\sim\!p^1_{\bx\bz}$ have unusually large $\|\bx-\bz\|_2$ (above the dashed 90th-percentile threshold).
We find that these long-distance pairs are also systematically more curved under our curvature proxy (Sec.~\ref{sec:curvature}), as shown in Fig.~\ref{fig:trunc1} (right).
Motivated by this correlation, we introduce a simple \emph{distance-based truncation} heuristic: we discard rectified couplings whose $\|\bx-\bz\|_2$ exceeds a fixed percentile threshold.
In practice, we found that truncating the top \(\mathbf{10\%}\) largest-distance couplings works robustly across all settings.

\subsubsection{Algorithm.} The full algorithm of \Approach{} is provided in Alg.\ref{alg:reMeanFlow}.

\section{Experiments}
\label{sec:experiment}

\subsubsection{Datasets.}
We conduct experiments on ImageNet~\cite{deng2009imagenet} at $64^2$ resolution in pixel space and at $256^2$ and $512^2$ resolutions in latent space~\cite{rombach2022high}.

\subsubsection{Settings.}
We initialize \Approach from a pretrained flow or diffusion model. For ImageNet-$64^2$ and ImageNet-$512^2$, we initialize \Approach from the pretrained EDM2-S model~\cite{karras2024analyzing}. For ImageNet-$256^2$, we initialize from the pretrained SiT-XL model~\cite{ma2024sit}. To rectify the trajectory, we generate 5M data-noise couplings using the default procedures described in the corresponding papers. During sampling for better synthetic image quality, we apply classifier-free guidance (CFG)\cite{ho2022classifier} for ImageNet-$256^2$ and Autoguidance\cite{karras2024guiding} for ImageNet-$64^2$ and $512^2$. More details on the implementation and the hyperparameter settings are provided in Appendix ~\ref{sec: hyper}.

\subsubsection{Baselines.}
We compare \Approach against recent state-of-the-art one-step flow-based methods, selecting baselines with comparable architecture or computational cost to ensure fair comparisons.
\begin{table*}[t]
    
    \centering
    \tiny
    \captionsetup{font=small}
    \setlength{\tabcolsep}{2.0pt}     
    \renewcommand{\arraystretch}{0.8} 
    \caption{\textbf{Class-conditional generation on ImageNet.} All results use classifier-free guidance (CFG) when supported; ``$\times 2$'' indicates that CFG doubles the effective NFE. $^\dagger$ denotes methods initialized from, or directly comparable to, the diffusion backbones also marked with $^\dagger$. (iCT\cite{song2023improved} result at $256^2$ is reported by IMM\cite{zhou2025inductive}, and ECT/ECD\cite{geng2024consistency} results at $512^2$ are reported by CMT\cite{hu2025cmt}). 
    \textbf{\Approach{} (ours) achieves the best FID across all settings, outperforming prior state-of-the-art one-step distillation and training-from-scratch methods.}}

    \resizebox{0.95\textwidth}{!}{%
    \begin{minipage}{\textwidth}
    \centering

    \begin{subtable}[t]{0.3\linewidth}
        \centering
        \caption{ImageNet $64^2$}
        \vspace{-6pt}
        \begin{tabular}{@{}lcc@{}}
            \toprule
            \textbf{Method} & \textbf{NFE} & \textbf{FID} \\
            \midrule
            \multicolumn{3}{@{}l}{\textbf{Diffusion models}}\vspace{1pt}\\
            EDM2-S~\cite{karras2024analyzing}$^\dagger$ & $63\times2$ & 1.58 \\
            \ \ + Autoguidance~\cite{karras2024guiding} & $63\times2$ & 1.01 \\
            EDM2-XL~\cite{karras2024analyzing} & $63\times2$ & 1.33 \\
            \midrule
            \multicolumn{3}{@{}l}{\textbf{Few-step models}}\vspace{1pt}\\
            2-rectified flow++~\cite{lee2024improving} & 1 & 4.31 \\
            iCT~\cite{song2023improved} & 1 & 4.02 \\
            ECD-S~\cite{geng2024consistency}$^\dagger$ & 1 & 3.30 \\
            sCD-S~\cite{lu2024simplifying}$^\dagger$ & 1 & 2.97 \\
            TCM~\cite{lee2024truncated}$^\dagger$ & 1 & 2.88 \\
            AYF~\cite{sabour2025align}$^\dagger$ & 1 & 2.98 \\
            \rowcolor{torange}
            \textbf{\Approach~(ours)}$^\dagger$ & \textbf{1} & \textbf{2.87} \\
            \bottomrule
        \end{tabular}
    \end{subtable}
    \hfill
    \begin{subtable}[t]{0.31\linewidth}
        \centering
        \caption{ImageNet $256^2$}
        \vspace{-6pt}
        \begin{tabular}{@{}lcc@{}}
            \toprule
            \textbf{Method} & \textbf{NFE} & \textbf{FID} \\
            \midrule
            \multicolumn{3}{@{}l}{\textbf{Diffusion models}}\vspace{4pt}\\
            ADM~\cite{dhariwal2021diffusion} & $250\times2$ & 10.94 \\
            DiT-XL~\cite{peebles2023scalable} & $250\times2$ & 2.27 \\
            SiT-XL~\cite{ma2024sit}$^\dagger$ & $250\times2$ & 2.06 \\
            SiT-XL + REPA~\cite{yu2024representation} & $250\times2$ & 1.42 \\
            \midrule
            \multicolumn{3}{@{}l}{\textbf{Few-step models}}\vspace{4.7pt}\\
            iCT~\cite{song2023improved} & 1 & 34.6 \\
            SM~\cite{frans2024one} & 1 & 10.6 \\
            iSM~\cite{nguyen2025improved} & 1 & 5.27 \\
            iMM~\cite{zhou2025inductive} & $1\times2$ & 7.77 \\
            MeanFlow~\cite{geng2025mean} & 1 & 3.43 \\
            \rowcolor{torange}
            \textbf{\Approach~(ours)}$^\dagger$ & \textbf{1} & \textbf{3.41} \\
            \bottomrule
        \end{tabular}
    \end{subtable}
    \hfill
    \begin{subtable}[t]{0.3\linewidth}
        \centering
        \caption{ImageNet $512^2$}
        \vspace{-6pt}
        \begin{tabular}{@{}lcc@{}}
            \toprule
            \textbf{Method} & \textbf{NFE} & \textbf{FID} \\
            \midrule
            \multicolumn{3}{@{}l}{\textbf{Diffusion models}}\\
            EDM2-S~\cite{karras2024analyzing}$^\dagger$ & $63\times2$ & 2.23 \\
            \ \ + Autoguidance~\cite{karras2024guiding} & $63\times2$ & 1.34 \\
            EDM2-XXL~\cite{karras2024analyzing} & $63\times2$ & 1.81 \\
            \midrule
            \multicolumn{3}{@{}l}{\textbf{Few-step models}}\\
            ECT$^\dagger$~\cite{geng2024consistency} & 1 & 9.98 \\
            ECD$^\dagger$~\cite{geng2024consistency} & 1 & 8.47 \\
            CMT~\cite{hu2025cmt}$^\dagger$ & 1 & 3.38 \\
            sCT-S~\cite{lu2024simplifying} & 1 & 10.13 \\
            sCD-S~\cite{lu2024simplifying}$^\dagger$ & 1 & 3.07 \\
            AYF~\cite{sabour2025align}$^\dagger$ & 1 & 3.32 \\
            \rowcolor{torange}
            \textbf{\Approach~(ours)}$^\dagger$ & \textbf{1} & \textbf{3.03} \\
            \bottomrule
        \end{tabular}
    \end{subtable}

    \end{minipage}%
    } 
    \label{tab: main}
\end{table*}

\begin{figure}[t]
\centering
\small
\begin{minipage}[t]{0.32\linewidth}
    \centering
    \includegraphics[width=\linewidth]{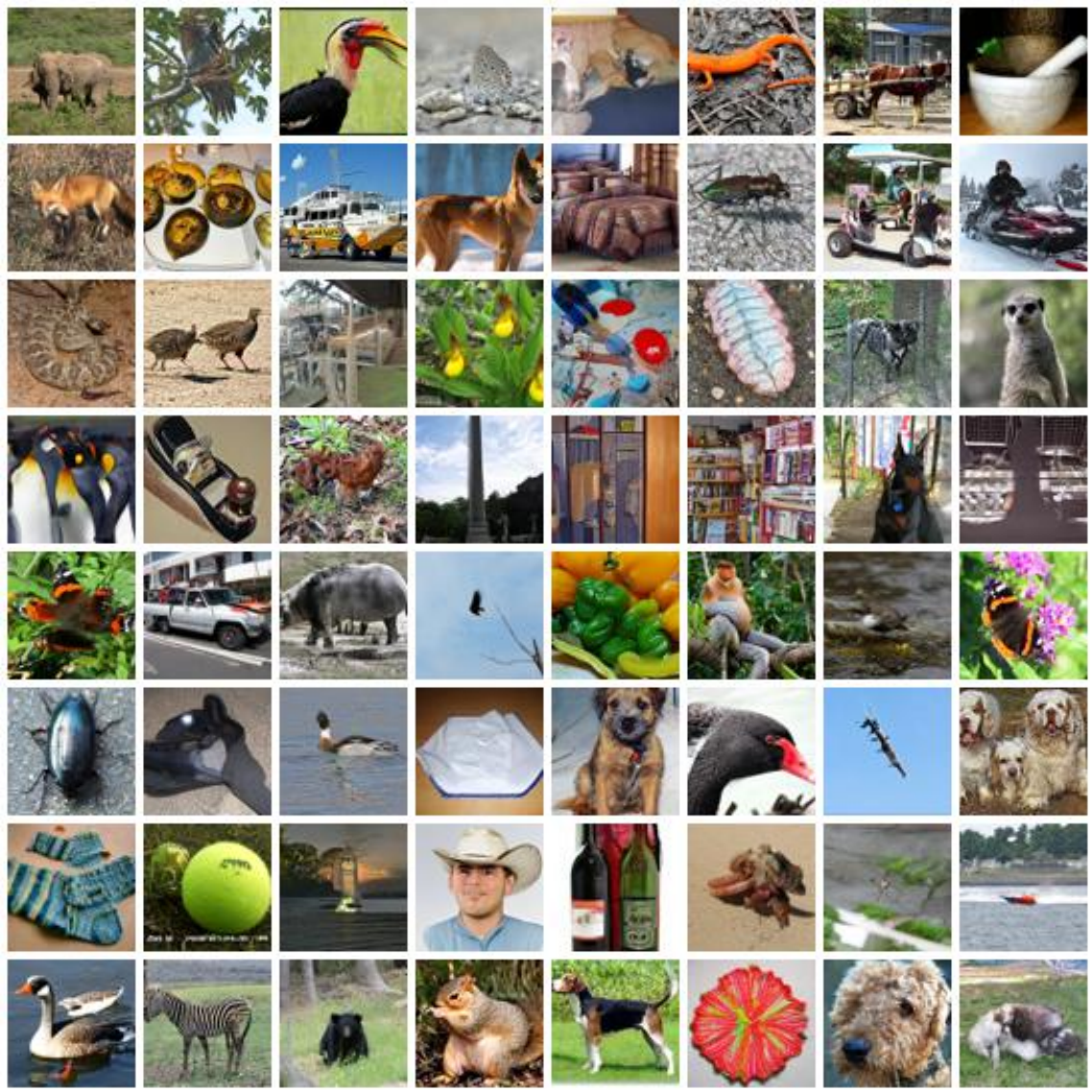}
\end{minipage}
\begin{minipage}[t]{0.32\linewidth}
    \centering
    \includegraphics[width=\linewidth]{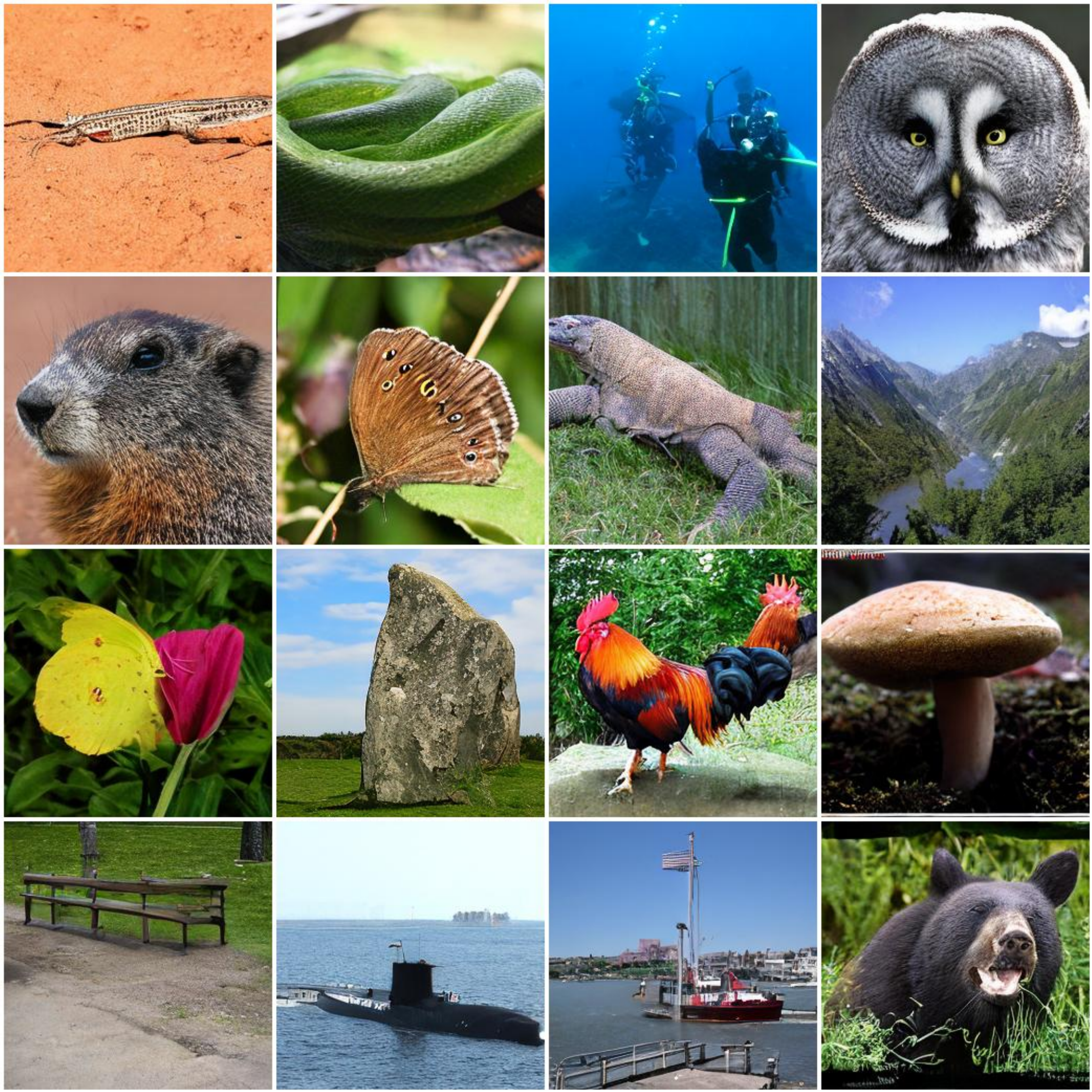}
\end{minipage}
\begin{minipage}[t]{0.32\linewidth}
    \centering
    \includegraphics[width=\linewidth]{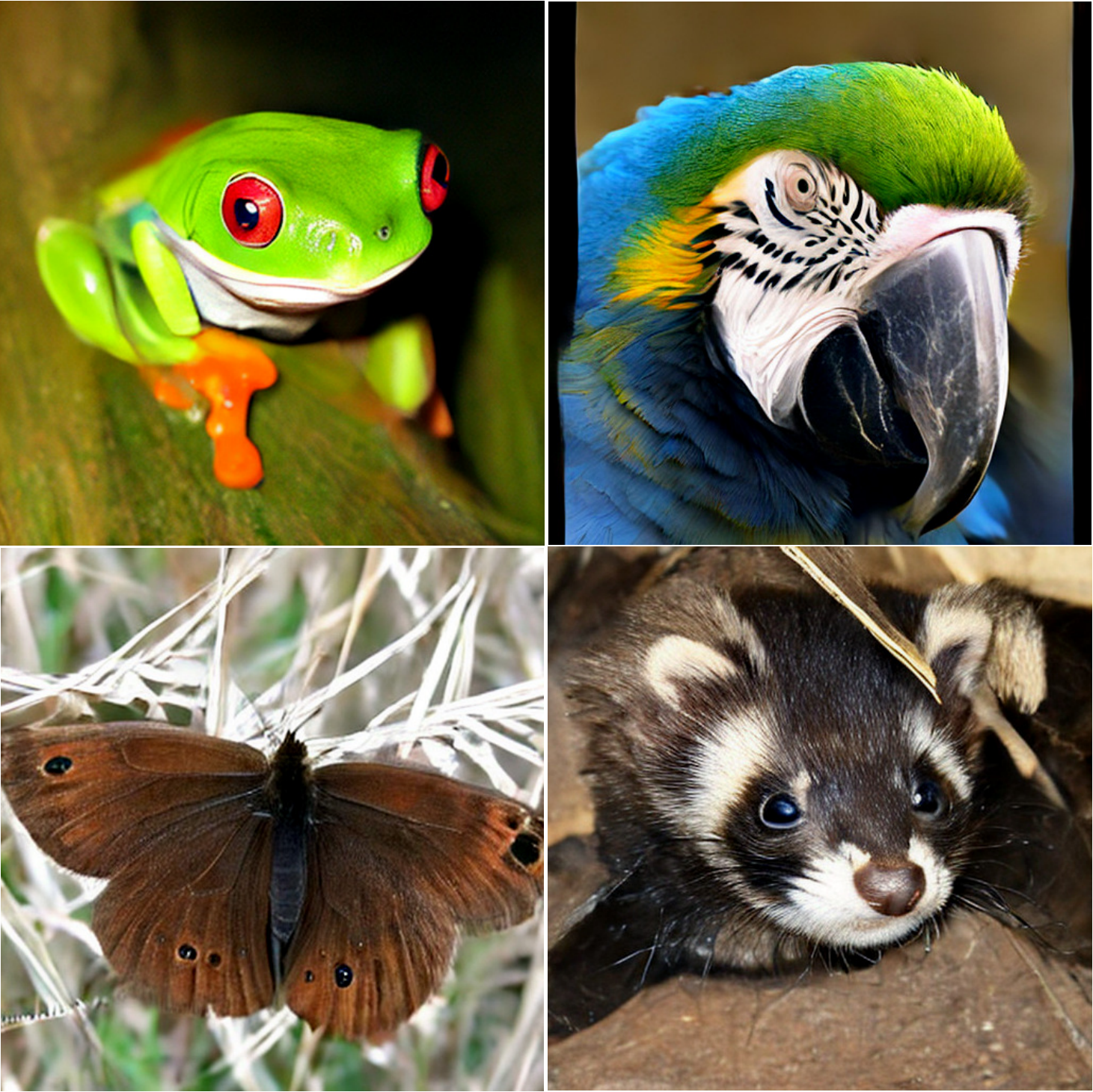}
\end{minipage}
\caption{\textbf{Visualization of \Approach{} one-step generation (NFE=1) on ImageNet at $64^2$ (Left), $256^2$ (Middle), and $512^2$ (Right).}}
\label{fig:qualitative-64-256}
\end{figure}

\subsection{One-Step Generation Quality}
For all experiments, we evaluate image quality using Fréchet Inception Distance (FID)~\cite{heusel2017gans} in Tab.\ref{tab: main}. When comparing with prior methods, we select those with comparable architectures and parameter counts. Across all evaluated resolutions, \textbf{\Approach consistently outperforms state-of-the-art one-step flow-based generation methods}. The qualitative results are shown in Fig.~\ref{fig:qualitative-64-256}

On the EDM2-S backbone~\cite{karras2024analyzing}, \Approach{} achieves superior performance at both $64^2$ and $512^2$ resolutions. On ImageNet-$64^2$, \Approach{} outperforms the closely related 2-rectified flow++~\cite{lee2024improving} by \textbf{33.4\%} in FID, and slightly improves over recent state-of-the-art one-step baselines\cite{lu2024simplifying, lee2024truncated, sabour2025align}. On ImageNet-$512^2$, \Approach{} also delivers strong one-step image quality, achieving a \textbf{9\%} FID gain over AYF\cite{sabour2025align} and outperforming strong consistency distillation methods\cite{sabour2025align, hu2025cmt}.

On the SiT backbone~\cite{ma2024sit} for ImageNet-$256^2$, \Approach{} slightly surpasses MeanFlow~\cite{geng2025mean} despite being trained \emph{without} real-image supervision and relying solely on synthesized samples. This is noteworthy because training exclusively on self-generated data is known to degrade performance due to self-conditioning effects~\cite{alemohammad2023self}. We attribute this improvement to the more favorable optimization landscape created by rectified mean-velocity learning: although the supervision is limited to synthetic couplings, rectification produces a smoother and lower-variance trajectory family. As a result, the model converges more reliably toward a high-quality solution.


\begin{figure}[t]
\includegraphics[width=1.\linewidth]{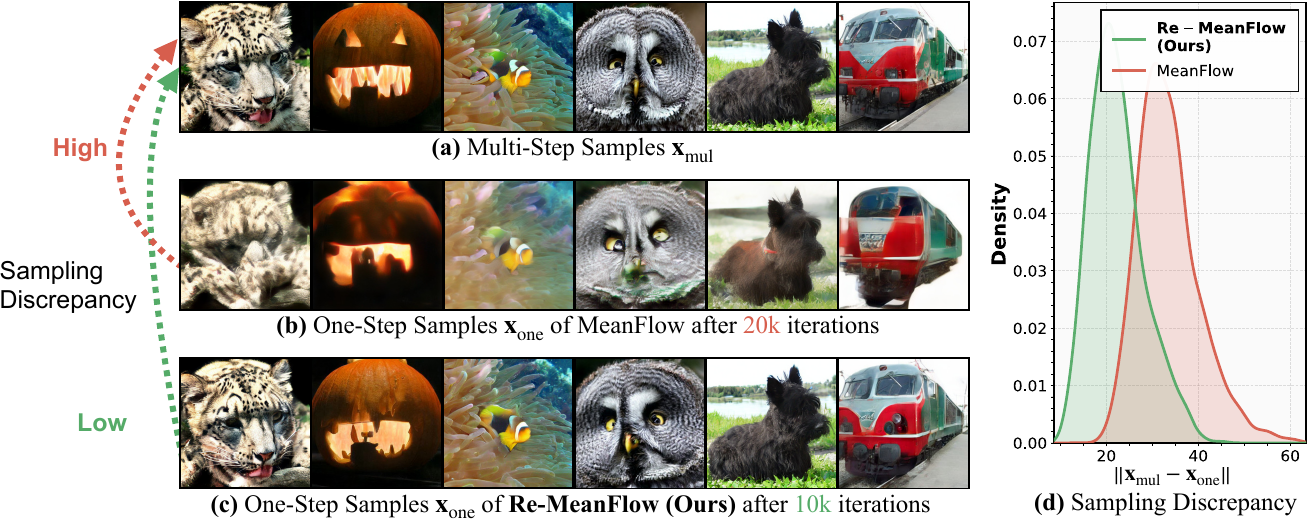}
\caption{\textbf{\Approach{} yields faithful one-step sampling with significantly less training.}
\textbf{(a)} Multi-step samples \(\bx_{\text{mul}}\) (reference) obtained by integrating the predicted mean-velocity field; \emph{multi-step results from MeanFlow and \Approach{} are visually similar, and we show a mixed set from both methods.}
\textbf{(b)} MeanFlow one-step samples \(\bx_{\text{one}}\) after 20k iterations remain blurry and deviate noticeably from \(\bx_{\text{mul}}\).
\textbf{(c)} \Approach{} one-step samples after 10k iterations are already sharp and closely match \(\bx_{\text{mul}}\).
\textbf{(d)} Sampling discrepancy distribution \(\|\bx_{\text{mul}}-\bx_{\text{one}}\|\): \Approach{} is strongly left-shifted, indicating substantially smaller one-step vs.\ multi-step mismatch.
\emph{All results are on ImageNet \(512^2\), using the \emph{same} noise seeds for both methods.}}
\label{fig: discrepancy}
\end{figure}

\subsection{Training Efficiency}
\label{sec: eff}

\subsubsection{Compared to MeanFlow.}We compare \Approach against MeanFlow on ImageNet $512^2$ \emph{without} classifier-free guidance (CFG). \textbf{As shown in Fig.~\ref{fig: main}, \Approach converges substantially faster:} even when MeanFlow is trained with $2\times$ the compute budget, its one-step samples remain noticeably blurry, whereas \Approach already produces sharp images, with a large FID gap (8.6 vs.\ 30.9).
To diagnose this difference, we contrast \emph{one-step} samples $\bx_{\mathrm {one}} = \bz - u_\theta(\bz, 0, 1) $  against \emph{multi-step} samples $\bx_{\mathrm {mul}} = \bz - \int_0^1 u_\theta (\bz, \tau, \tau) d\tau $ obtained by simulating the integral of the instantaneous-velocity (Fig.~\ref{fig: discrepancy}).
Both methods are initialized from the same EDM2-S checkpoint, while MeanFlow is trained with $2\times$ the compute budget.
Qualitatively, \Approach{} produces nearly indistinguishable samples under one-step and multi-step generation, indicating that $u_\theta(\bz,0,1)$ closely matches the multi-step estimate. In contrast, MeanFlow's one-step predictions are significantly degraded.
Quantitatively, we measure the sampling discrepancy as
\(
\|\bx_{\mathrm{mul}}-\bx_{\mathrm{one}}\|
\) and plot its distribution in Fig.~\ref{fig: discrepancy}d. \Approach exhibits a pronounced left shift, confirming a smaller one-step-to-multi-step error and hence more accurate mean-velocity modeling of the underlying trajectories. We provide additional convergence comparisons with MeanFlow and other related methods in Appendix~\ref{sec:convergence_ap}.

\subsubsection{Compared to Other One-Step Distillation Methods.}
As noted by \cite{lee2024improving}, distillation methods involving sampling rectified couplings can still remain computationally competitive in overall computation cost, even though they require generating additional data-noise couplings.

\begin{figure*}[t]
\centering
\small
\includegraphics[width=1.0\linewidth]{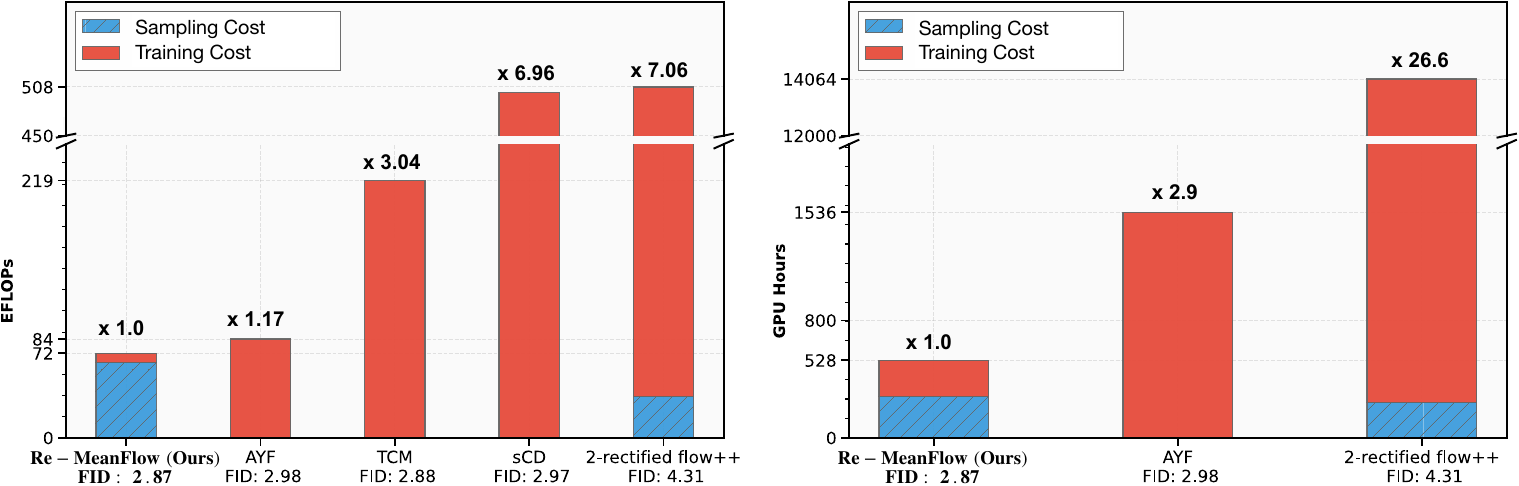}
\caption{\textbf{Total compute comparison on ImageNet-$64^2$.}
We compare against the strongest prior methods in our main FID table (Tab.~\ref{tab: main}) in terms of the quality-efficiency trade-off.
Total cost is reported in EFLOPs (left) and GPU hours (right), and each bar is decomposed into training (solid red) and coupling sampling (blue hatched).
\Approach{} is the baseline (1.0$\times$; numbers denote factors relative to \Approach{}).
Across both metrics, \textbf{\Approach{} achieves the lowest end-to-end compute, significantly outperforming prior approaches even after accounting for coupling sampling.}}
\label{fig:eff}

\end{figure*}

We evaluate \Approach by estimating the total computation cost in FLOPs and GPU hours on ImageNet-$64^2$. The protocol for computing these metrics follows \cite{lee2024improving} and is detailed in the Appendix~\ref{sec: flop}. We compare against AYF~\cite{sabour2025align}, the strongest existing distillation baseline in both quality and efficiency, and against the closely related 2-rectified flow++~\cite{lee2024improving}.

As shown in Fig.~\ref{fig:eff}, \textbf{\Approach achieves the lowest overall compute cost among recent distillation approaches.} In terms of GPU hours, \Approach is \textbf{26.6$\times$ faster} than 2-rectified flow++, reinforcing the results observed in our controlled efficiency experiments. Even compared to AYF, which does not require coupling generation, \Approach remains \textbf{2.9$\times$ faster}.

We also compare estimated FLOPs. Although \Approach{} shows a smaller advantage under this metric, FLOPs alone do not fully reflect practical runtime. A substantial portion of our compute lies in the inference-only coupling sampling stage, which can be executed on widely accessible inference-grade GPUs and runs significantly faster than training workloads with the same FLOP count. 

\begin{figure}[t]
    \includegraphics[width=1.\linewidth]{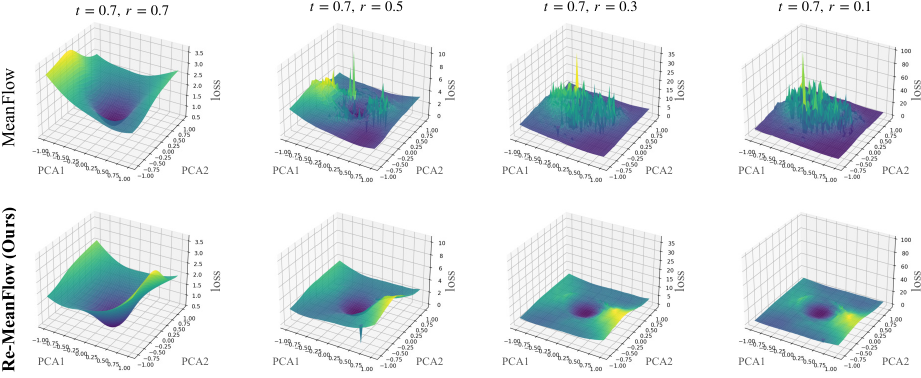}
    \caption{\textbf{Loss landscape of \(\mathcal{L}_{MF}\) on a PCA plane of \(\bz_t\).}
    We visualize $\mathcal{L}_{MF}$ as a function of the input $\bz_t$ by evaluating the loss on a 2D grid spanned by the top two PCA directions of a batch of $\bz_t$ samples. We fix $t=0.7$ and sweep $r\in\{0.7,0.5,0.3,0.1\}$ (left to right), where larger $t-r$ corresponds to a harder, longer-jump $(r,t)$ regression task.
    \textbf{Top (MeanFlow):} the landscape becomes increasingly sharp and irregular as $t-r$ grows.
    \textbf{Bottom (\Approach{}; ours):} the landscape remains substantially smoother and better-conditioned across all $r$.
    \textbf{Overall, learning mean velocity on straighter trajectories yields a markedly smoother objective, enabling more efficient and stable training.}}
    \label{fig:loss-landscape}

\end{figure}

\subsection{Loss Landscape}
\label{sec:loss-landscape}
To explain the training speedup enabled by learning mean-velocity on straighter trajectories, we visualize in Fig.~\ref{fig:loss-landscape} the loss landscape of $\mathcal{L}_{MF}$ on ImageNet $256^2$ with respect to the input $\bz_t$ for both MeanFlow and \Approach{}. We fix $t=0.7$ and plot the landscape for different values of $r$. Additional results at other resolutions are provided in Appendix~\ref{sec:loss_ab}.

To construct these visualizations, we perform PCA on a batch of $\bz_t$ samples to obtain the top two principal directions. We then evaluate $\mathcal{L}_{MF}$ on a 2D grid spanned by these directions and centered at the sampled $\bz_t$, and visualize the resulting surfaces.
As shown in Fig.~\ref{fig:loss-landscape}, \Approach{} consistently exhibits a smoother and better-conditioned landscape. \textbf{In particular, as $t-r$ grows, the MeanFlow objective becomes increasingly sharp and irregular, whereas \Approach{} remains substantially smoother and more stable.}

Unlike conventional loss-landscape studies \cite{li2018visualizing} that perturb model parameters, \textbf{we perturb $\bz_t$ to directly measure the conditioning of the mean-velocity regression induced by the trajectory geometry.} When trajectories are highly curved, the interval mean velocity becomes a rapidly varying, locally unstable function of $\bz_t$, yielding a rugged loss landscape in $\bz_t$-space. Such ruggedness indicates that the regression target itself is poorly conditioned, which in turn hampers optimization. In contrast, rectified (straighter) trajectories markedly smooth the landscape as a function of $\bz_t$, producing a better-conditioned learning problem and enabling more efficient training.

\begin{table}[t]
\centering
\scriptsize
\setlength{\tabcolsep}{3pt}
\captionsetup{font=small}
\caption{Effect of distance-based truncation strength (discarding a top fraction of couplings by $\ell_2$ endpoint distance). We report FID/IS and Precision/Recall on ImageNet $512^2$ (top) and $256^2$ (bottom). \textbf{Mild truncation consistently improves stability and sample quality} (lower FID, higher IS) without sacrificing diversity (comparable Precision/Recall). \textbf{Notably, even without truncation, \Approach{} matches MeanFlow trained from scratch on real data.}}
\label{tab:trunc}
\begin{tabular}{lcccc}
\toprule
\multicolumn{5}{c}{\textbf{ImageNet $512^2$}} \\
\midrule
\textbf{Method}  & \textbf{FID $\downarrow$} & \textbf{IS $\uparrow$} & \textbf{Precision $\uparrow$} & \textbf{Recall $\uparrow$} \\
\midrule
\textbf{\Approach} (no trunc.) & 3.50 & 242.62 & 0.73 & 0.54 \\
\textbf{\Approach} (top $5\%$) & 3.10 & 251.84 & 0.74 & \textbf{0.56} \\
\rowcolor{torange}
\textbf{\Approach} (top $10\%$) & \textbf{3.03} & \textbf{262.37} & \textbf{0.75} & 0.54 \\
\textbf{\Approach} (top $15\%$) & 3.19 & 259.61 & 0.76 & 0.53 \\
\midrule
\multicolumn{5}{c}{\textbf{ImageNet $256^2$}} \\
\midrule
MeanFlow~\cite{geng2025mean} & 3.43 & 247.5 & 0.73 & 0.54 \\
\textbf{\Approach} (ours; no trunc.) & 3.48 & 243.4 & 0.72 & 0.54 \\
\rowcolor{torange}
\textbf{\Approach} (ours; top $10\%$) & \textbf{3.41} & \textbf{249.6} & \textbf{0.73} & \textbf{0.54} \\
\bottomrule
\end{tabular}
\end{table}

\subsection{Distance-based Truncation}
As discussed in Sec.~\ref{sec:trunc1}, we empirically observe that large endpoint distances $\|\bz-\bx\|_2$ are correlated with higher trajectory curvature. We therefore apply a simple top-$10\%$ truncation rule, discarding the couplings with the largest $\ell_2$ endpoint distance. In practice, this filter removes many residual high-curvature pairs, improving training stability and sample quality for \Approach{}.

To study the effect of the truncation ratio, we sweep different truncation rates and report results on ImageNet $512^2$ and ImageNet $256^2$ (Tab.~\ref{tab:trunc}). We find that moderate truncation consistently improves stability and yields better IS/FID, while maintaining diversity as reflected by comparable Recall. In particular, FID improves monotonically within a moderate truncation regime.

Interestingly, even without truncation, \Approach{} already matches the performance of MeanFlow trained directly on real data. This is notable because \Approach{} does not require access to the original dataset: it learns solely from generated coupling pairs. Since self-generated distillation typically incurs discretization error that can degrade performance, we hypothesize that the rectified trajectories substantially simplify the optimization problem, enabling more effective parameter learning and thus competitive (or improved) results.

\label{sec:ablation}
\subsection{Ablation Study}

\begin{table}[t]        
\centering
\small
\caption{Ablation study on ImageNet-$512^2$ for key implementation details.}
\label{tab:uncond-fid-only}
\begin{tabular}{lc}
\toprule
Training configurations & FID $\downarrow$ \\
\midrule
Base (Best Setting reported in \cite{geng2025mean}) & 7.81 \\
(a) + Hyperparameter adjustments & 7.22 \\
(b) + Time embedding change & 4.60 \\
(c) + U-shaped $t$ distribution & 3.71 \\
(d) + Avoid high-variance $(r,t)$ region & 3.50 \\
\rowcolor{torange}
(e) + Distance-based truncation & 3.03 \\
\bottomrule
\end{tabular}
\label{tab:ablation}
\end{table}

To identify the implementation choices most critical for stable and high-quality training, we conduct an ablation study on ImageNet-$512^2$ in Tab.~\ref{tab:ablation}. We summarize the main findings below.

\noindent\textbf{(a) Hyperparameter adjustments.}
Since rectified trajectories are substantially straighter, we reduce the normalization strength in MeanFlow's adaptive loss from $1.0$ to $0.5$ (Pseudo-Huber style).

\noindent\textbf{(b) Time embedding change.}
Following~\cite{sabour2025align, lu2024simplifying}, we replace the EDM2 time embedding $\mathrm{emb}(\log\sigma_t)$ with $\mathrm{emb}(t)$ to improve the stability of Jacobian-vector product computations.

\noindent\textbf{(c) U-shaped $t$ distribution.}
Because rectified trajectories reduce the need for mid-range emphasis, we adopt the U-shaped $t$ distribution from~\cite{lee2024improving} in place of the standard lognormal schedule.

\noindent\textbf{(d) Avoid high-variance $(r,t)$ region.}
We observe unusually high variance when $t>0.95$ and $r<0.4$ (analyzed in the Appendix ~\ref{sec: var}). Inspired by the truncation strategy in TCM~\cite{lee2024truncated}, we exclude this time region, which improves FID and accelerates convergence.

\noindent\textbf{(e) Distance-based truncation.}
Finally, applying our distance-based truncation heuristic (Sec.~\ref{sec:trunc1}) further improves both efficiency and generation quality.

\section{Conclusion}
We introduced \textbf{Re}ctified-\textbf{MeanFlow}, a lightweight, \emph{data-free} self-distillation framework for one-step generation that addresses a key MeanFlow bottleneck: learning mean-velocity on highly \emph{curved} trajectories. Our central geometric takeaway is that mean-velocity estimation is substantially easier on \emph{straighter} paths. Accordingly, \Approach{} models mean-velocity on \emph{rectified} couplings generated from a pretrained flow model, yielding a simpler velocity field and a markedly better-conditioned loss landscape; this translates into faster convergence and stronger one-step generation. We further improve robustness with a distance-based truncation rule that removes residual high-curvature couplings. Empirically, on ImageNet at $64^2$, $256^2$, and $512^2$, \Approach{} consistently outperforms prior one-step flow distillation methods and strong Rectified Flow baselines, achieving higher sample quality with substantially less training.

\bibliographystyle{splncs04}
\bibliography{main}

\clearpage
\appendix
\renewcommand{\theHsection}{appendix.\Alph{section}}
\counterwithout{table}{section}
\setcounter{table}{3}
\setcounter{figure}{7}
\setcounter{page}{1}
\startcontents[sections]

\section*{Appendix}
\begingroup
    \hypersetup{linkcolor=toclink}
    \setcounter{tocdepth}{2}
    \def\authcount#1{}
    \printcontents[sections]{l}{1}{}
\endgroup

\section{More Discussion}

In this section, we provide additional analysis and context for our method. We begin by discussing its limitations and then examine the broader impact of our approach. Next, we revisit the theoretical motivation for using a single rectification iteration and explain how our perspective complements existing analyses. Finally, we compare our method with the concurrent CMT method~\cite{hu2025cmt}.

\subsection{Limitation}
Since our method does not access real data during training and relies entirely on synthetic samples generated by pretrained diffusion or flow models, its performance naturally depends on the quality of these generated couplings. A promising direction is to incorporate real data into the training process, as explored in \cite{lee2024improving, seong2025balanced}. Another complementary line of work investigates how to improve generative models using only their own outputs~\cite{alemohammad2023self, alemohammad2024self, alemohammad2025neon}, which suggests that self-generated data can still be leveraged effectively with appropriate regularization. Alternatively, one could improve the synthetic supervision directly by using stronger backbone models to generate the couplings.

\subsection{Broader Impact}

Beyond empirical gains, our method highlights a practical and socially relevant shift in how large generative models can be trained. Traditional few-step or one-step distillation pipelines often concentrate compute in expensive training workloads on high-end accelerators (e.g., A100-class GPUs), which limits accessibility to well-resourced institutions. In contrast, \Approach{} shifts a substantial fraction of computation to an \emph{inference-only} stage (sampling rectified couplings), followed by a lightweight MeanFlow training phase. Because inference workloads can be executed efficiently on widely available consumer- or inference-grade accelerators, our framework reduces reliance on scarce training GPUs and lowers the barrier to experimentation.

Practically, \textbf{the sampling process for the rectified couplings is amenable to pipeline parallelism}: coupling generation can run asynchronously alongside training as a data-preparation stream. For implementation convenience, we pre-generate couplings before training in this work; nevertheless, even under this conservative setup,\textbf{ \Approach{} achieves the lowest total compute among prior state-of-the-art distillation methods}, underscoring the end-to-end efficiency of our framework.

\subsection{Comparison with One-Step Sampling via Rectified Flow}
\label{sec:reflow-once-summary}
Rectified Flow~\cite{liu2022flow} supports few-step sampling via an iterative \emph{reflow} procedure that progressively straightens trajectories by repeatedly training a new flow model on rectified couplings. Liu et al.~\cite{liu2022flow} show that, in the limit, iterating this procedure can yield perfectly straight couplings. In practice, however, \textbf{achieving near-linear paths for reliable one-step sampling typically requires multiple reflow rounds}, which is computationally expensive and can degrade performance due to accumulated approximation errors.

Lee et al.~\cite{lee2024improving} analyze when multiple Reflow iterations are truly necessary for achieving straight trajectories in rectified flows. Their argument centers on how trajectory \emph{intersections} affect the learned velocity field. 
Consider two 1-rectified couplings $(\x',\z')$ and $(\x'',\z'')$. A trajectory intersection occurs if there exists $t\in[0,1]$ such that
\[
(1-t)\x' + t\z' = (1-t)\x'' + t\z'' .
\]
If such an intersection happens, both trajectories pass through the same intermediate point. Because rectified-flow training regresses the conditional expectation
$E[\x | \x_t]$, the model must assign a \emph{single} velocity to this shared point. As a result, the learned velocity field cannot simultaneously point toward both $\x'$ and $\x''$, and it instead averages their directions. This averaging effect bends the local velocity field, producing curvature and consequently degrading the accuracy of one-step Euler sampling, which assumes the path to be straight.

To understand how often this phenomenon can happen, Lee et al.~\cite{lee2024improving} show that an intersection implies
\[
\z'' = \z' + \frac{1-t}{t}(\x' - \x'').
\]
Under typical training, nearly all noise samples used to form 1-rectified couplings lie in the high-density region of the Gaussian prior. The $\z''$ required above usually lies far outside that region unless $||\x'-\x''||_2$ is extremely small or $t$ is very close to~1. Therefore, intersections are statistically rare.
Further assuming that the 1-rectified flow is approximately $L$-Lipschitz,
\begin{align}
    ||\x' - \x''||_2 \le L\, ||\z' - \z''||_2,
    \label{eq:lip}
\end{align}
nearby noise samples cannot map to widely separated data points. Combining the rarity of intersections with this Lipschitz condition, the authors conclude that the optimal 2-rectified flow is nearly straight. Hence, in their view, one additional Reflow step is sufficient, and any remaining performance gap should be attributed primarily to training inefficiency rather than insufficient straightening.

\subsubsection{Relation to Our Work.}
Our focus is complementary. While the above analysis suggests that trajectory intersections are rare for most couplings, our experiments indicate that realistic settings can still contain a small but influential subset of pairs with non-negligible curvature, particularly when the effective Lipschitz constant $L$ is large due to geometric imbalance in the data distribution. As shown in Fig.~\ref{fig:method}b and Fig.~\ref{fig: curv}, trajectories induced by a once-rectified coupling (i.e., one reflow step) are substantially straighter than those from the independent coupling (Fig.~\ref{fig:method}a), yet residual curvature remains and can still hinder reliable one-step sampling based on the instantaneous velocity. \Approach{} is designed to be robust in precisely these challenging cases, enabling stable one-step generation even when curvature persists after a single rectification step.

\subsection{Comparison with CMT}
CMT~\cite{hu2025cmt} is an important concurrent effort that also leverages synthetic trajectories generated by a pretrained sampler to stabilize few-step flow-map training. Conceptually, CMT introduces a dedicated mid-training stage that learns a full trajectory-to-endpoint mapping from solver-generated paths, which then serves as a trajectory-aligned initialization for a subsequent post-training flow-map stage. In contrast, our method adopts a fundamentally different design: rather than supervising on entire solver trajectories, we distill only the end-point couplings of rectified flows and learn the corresponding mean velocity in a single training stage. This distinction yields a practical advantage: our pipeline avoids the compounded complexity of CMT’s two-stage optimization, which is sensitive to hyperparameters at both stages and substantially more expensive to tune.

\section{Implementation Details}
\label{sec: hyper}
In this section, we first describe the conditioning strategy of how \Approach utilizes previous pretrained models. We then outline the key design choices during training \Approach. We provide the training hyperparameters across all ImageNet resolutions in Tab.~\ref{tab:hyper}.

\begin{table}[t]
    \centering
    \caption{\small Training settings of \Approach on ImageNet.}
    \label{tab:appendix:setting_in64}
    \setlength{\tabcolsep}{3pt}
    \renewcommand{\arraystretch}{0.96}
    \small
    \resizebox{\linewidth}{!}{%
    \begin{tabular}{|l|ccc|} %
        \hline %
         & \multicolumn{3}{c|}{Resolution} \\
         & $64^2$ & $256^2$ & $512^2$ \\
        \hline
        \multicolumn{1}{|l@{}}{\textbf{Training Details}} & & & \\
        \hline
        Model Backbone & EDM2-S\cite{karras2024analyzing} & SIT-XL\cite{ma2024sit} & EDM2-S\cite{karras2024analyzing} \\
        Global Batch size & 128 & 128 & 128 \\ 
        Learning Rate & 1e-4 & 1e-4 & 1e-4 \\
        Adam $\beta_1$ & 0.9 & 0.9 & 0.9 \\
        Adam $\beta_2$ & 0.99 & 0.95 & 0.99 \\
        Model Capacity (Mparams) & 280.2 & 676.7 & 280.5 \\ 
        EMA Rate & 0.9999 & 0.9999 & 0.9999 \\
        \hline
        \multicolumn{1}{|l@{}}{\textbf{MeanFlow Setting Details}} & & & \\
        \hline
        Ratio of $r\neq t$ & 0.25 &  0.25 &  0.25 \\
        $p$ for Adaptive Weight & 0.5 &  0.5 &  0.5 \\
        CFG Effective Scale $w'$ & Uniform(1.0, 3.0) &  Uniform(1.0, 3.0) &  Uniform(1.0, 2.5) \\
        Avoiding High-variance $(t,r)$ Region & $t>0.95 \ \& \ r<0.4$ & $t>0.95 \ \& \ r<0.4$ &  $t>0.95 \ \& \ r<0.4$ \\
        
        \hline
        \multicolumn{2}{|l@{}}{\textbf{Sampling Details for Rectified Couplings }} & &  \\ \hline
        Pretrained Model & EDM2-S\cite{karras2024analyzing} & SIT-XL\cite{ma2024sit} & EDM2-S\cite{karras2024analyzing} \\
        Sampling Number & 5M & 5M & 5M \\
        Guidance Method & Autoguidance \cite{karras2024guiding} & CFG \cite{ho2022classifier} & Autoguidance \cite{karras2024guiding} \\
        Distance Truncate Strength & Top 10\% & Top 10\% & Top 10\% \\
        \hline
    \end{tabular}%
    }
    \label{tab:hyper}
\end{table}

\subsection{Pretrained Model Conditioning}
In \Approach, the velocity network is conditioned on two time variables, $t$ and $r$. We implement this conditioning by learning two separate embeddings, $\mathrm{emb}_t(t)$ and $\mathrm{emb}_r(r)$, and summing them before passing the result to the rest of the network.
When initializing from pretrained models, the original networks only contain a single time embedding for $t$. Replacing this embedding with our two-embedding design requires careful initialization to ensure the model initially behaves like the pretrained flow mode. Specifically, before MeanFlow fine-tuning, we want:
\begin{equation}
    u(\x_t,t,r)\approx v(\x_t,t)
\end{equation}

\subsubsection{On ImageNet $64^2$ and $512^2$}, which \Approach is initialized from the pretrained EDM2-S model \cite{karras2024analyzing}, following AYF \cite{sabour2025align}, we perform a short alignment stage in which we train the new embeddings to reproduce the original time-embedding output for the corresponding noise level. Specifically, for EDM2-S the original embedding depends on $\log \sigma_t$, where $\sigma_t = \tfrac{t}{1-t}$. We train the new embeddings via:
\begin{equation}
\mathbb{E}_{t,r}[||\mathrm{emb}_t(t)+\mathrm{emb}_r(r)-\mathrm{emb}_{\mathrm{ori}}(\log \sigma_t)||_2^2],
\end{equation}
for 10k iterations with learning rate 1e-3. This process takes only a few minutes. We also convert the original VE diffusion parameterization into the flow-matching setting following \cite{lee2024improving}.

\subsubsection{On ImageNet $256^2$}, we initialize \Approach from SiT-XL~\cite{ma2024sit}, which is already a flow-based model. In this case, only the additional r-embedding needs to be introduced. We simply zero-initialize $\mathrm{emb}_r(r)$ and keep the original SiT time embedding for $\mathrm{emb}_t(t)$.

\subsection{Time Distribution}
We observe a similar loss--time profile to that reported in \cite{lee2024improving}: the training loss as a function of $t$ closely matches that of 2-rectified flow++. Accordingly, for sampling $t$ we adopt the same U-shaped distribution:
\[
p_t(u) \propto \exp(au) + \exp(-au),\quad u\in[0,1],\ a=4.
\]

Following AYF~\cite{sabour2025align}, after sampling $t$ we draw the interval length $|t-r|$ from a normal distribution $\mathcal{N}(P_{\text{mean}}, P_{\text{std}})$ and apply a sigmoid transformation. We use the same parameters as AYF, $(P_{\text{mean}}, P_{\text{std}})=(-0.8, 1.0)$, which emphasize medium-length intervals and substantially improve stability. As shown in Table~\ref{tab:uncond-fid-only}, this setting yields a noticeable improvement in FID and accelerates convergence.

We also experimented with sampling $t$ uniformly. Despite its simplicity, uniform sampling performs competitively---and in many cases better---than commonly used log-normal time distributions in diffusion and flow-matching models~\cite{karras2022elucidating, karras2024analyzing, geng2025mean}, which prioritize mid-range timesteps to avoid high variance near $t\approx 0$ and $t\approx 1$. We attribute the strong performance of uniform sampling to the significantly lower variance of rectified trajectories: after one reflow step, early-time and high-noise regions become much more stable, allowing us to allocate more samples to these challenging regimes without the usual degradation observed when training on independent couplings.

\begin{figure}[t]
    \centering
    \includegraphics[width=0.8\linewidth]{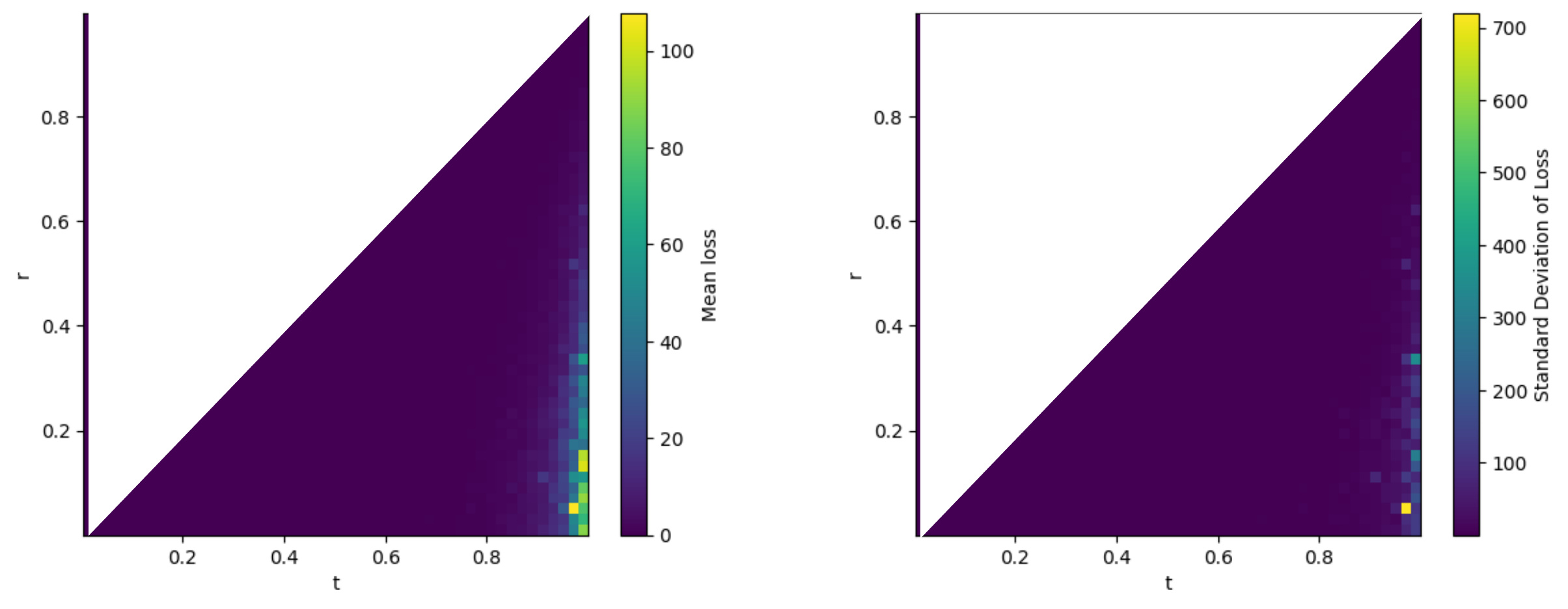}
    \caption{\textbf{Heatmaps of the mean loss (left) and the standard deviation of the loss (right)} for a trained \Approach model on ImageNet-$512^2$ under configuration (c).}
    \label{fig: high-var-tr}
    \vspace{-5mm}
\end{figure}
\subsection{Avoiding High-Variance Time Regions}
\label{sec: var}
As discussed in Sec.~\ref{sec:ablation}, we observe that \Approach exhibits unusually high variance when the noise level $t$ is large while the reference time $r$ is close to zero. Intuitively, this corresponds to asking the model to predict a \emph{slightly denoised} sample ($r \in [0,0.4]$) from an input that remains heavily corrupted ($t \approx 1$). To quantify this effect, we sample 100k pairs of $(t,r)$ uniformly and evaluate a trained \Approach model under configuration (c) in Table~\ref{tab:uncond-fid-only}. As shown in Fig.~\ref{fig: high-var-tr}, the resulting loss landscape displays a clear spike in both error and variance within this region, often even higher than the loss incurred when predicting directly from noise to a clean target.

Inspired by the truncation strategy in TCM~\cite{lee2024truncated}, we adopt a simple yet effective rule to avoid this problematic regime: whenever a sampled pair satisfies $t > 0.95$ and $r < 0.4$, we set $r = 0$. Empirically, this improves both training stability and FID. Our hypothesis is that predicting a \emph{clean} image from pure noise ($r = 0$, $t\approx1$) is substantially easier than predicting a lightly corrupted target: the latter requires the model to determine which noise components should be preserved, introducing ambiguity and variance at high $t$. By redirecting training toward these easier high-$t$ targets, the model can allocate more capacity to learning accurate one-step predictions. This modification not only improves FID relative to configuration (c), but also accelerates convergence: in the high-$t$ regime, more updates involve $r = 0$, allowing the model to refine its one-step outputs more quickly and reliably.

\subsection{Training with Guidance}
Classifier-free guidance (CFG)~\cite{ho2022classifier} is widely used to boost the performance of diffusion and flow-based generative models. To incorporate CFG into the MeanFlow stage, we train \Approach on the CFG-enhanced velocity field:
\begin{equation}
    v^{\text{cfg}}(\mathbf z_t, t \mid c) \triangleq \omega\, v(\mathbf z_t, t \mid c) + (1 - \omega)\, v(\mathbf z_t, t).
\end{equation}
MeanFlow~\cite{geng2025mean} further introduced an improved CFG method that mixes conditional and unconditional mean-velocity predictions:
\begin{equation}
    v^{\text{cfg}}(\mathbf z_t, t \mid c) = \omega\, v(\mathbf z_t, t \mid c) + \kappa u^{\text{cfg}}(\mathbf z_t,t,t|c) + (1 - \omega + \kappa)\, u^{\text{cfg}}(\mathbf z_t,t,t).
\end{equation}
which is equivalent to using an effective guidance of $\omega' = \frac{\omega}{1-\kappa}$. We adopt this improved CFG formulation for all experiments. 

Empirically, we found that directly training MeanFlow on the CFG field is unstable, consistent with observations in \cite{hu2025cmt}. To mitigate this, we use a simple two-stage strategy: first train $u_\theta$ on the unconditional flow, then on the CFG-modified flow. Usually, allocating half of the total training budget to each stage provides a good balance between stability and final quality.

We also found that sampling a random CFG scale $\omega'$ from a uniform distribution (rather than fixing it) gives better results. Large values of $\kappa$ are also important for stable training. In practice, we sample $\omega'$ from a uniform distribution, then set $\kappa = \max(1.0, \omega'-1)$, and finally compute the corresponding value for $\omega = \frac{\omega'}{1-\kappa}$.

\begin{figure*}[t]
    \centering
    \vspace{-5pt}
    \includegraphics[width=1.\linewidth]{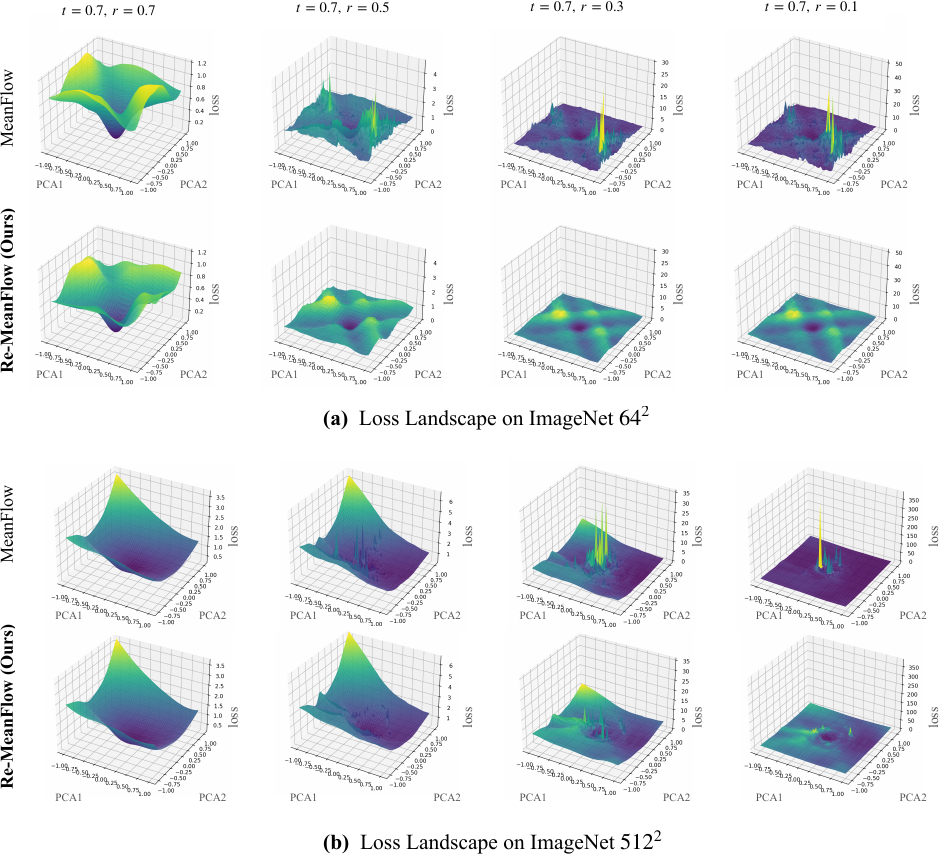}
    \caption{\textbf{Loss landscape of $\mathcal{L}_{MF}$ on a PCA plane of $\bz_t$} for ImageNet $64^2$ \textbf{(a)} and $256^2$ \textbf{(b)}. We evaluate $\mathcal{L}_{MF}$ on a 2D PCA grid with $t=0.7$ and $r\in\{0.7,0.5,0.3,0.1\}$ (left to right). \textbf{MeanFlow (top)} becomes increasingly sharp and irregular as $t-r$ grows, whereas \textbf{\Approach{} (bottom) remains substantially smoother and better-conditioned, supporting more stable and efficient training.}}
    \label{fig: loss2}
    \vspace{-15pt}
\end{figure*}

\section{More Experiment Details and Results }
In this section, we provide additional experimental details, computational analyses, and extended results. We first present loss-landscape visualizations at $64^2$ and $256^2$, then describe our protocol for estimating FLOPs and GPU-hours across all methods. Finally, we report additional convergence results that further support the synergy between trajectory rectification and mean-velocity modeling.

\subsection{Loss Landscape}
\label{sec:loss_ab}
In Fig.~\ref{fig:loss-landscape}, we visualize the loss landscape of $\mathcal{L}_{MF}$ with respect to the input $\bz_t$ for MeanFlow and \Approach{} across different $(t,r)$ pairs. Here we report the corresponding results at additional resolutions in Fig.~\ref{fig: loss2}, following the same protocol.

Concretely, we fix $t=0.7$ and sweep $r$, and compute PCA directions by collecting $20\,$k samples of $\bz_t$ (equivalently, $\bx_t$ in our implementation). For each resolution, we then evaluate $\mathcal{L}_{MF}$ on a 2D grid spanned by the top two PCA directions using a minibatch of samples. Collecting the data and generating all surfaces for a single resolution takes approximately $3$ hours on one A100 GPU.
Overall, consistent with Fig.~\ref{fig:loss-landscape}, \textbf{\Approach{} exhibits a substantially smoother and better-conditioned loss landscape, which translates into more efficient and stable training.}

\subsection{Computation Estimation of Each Method}
\label{sec: flop}
\subsubsection{FLOPS Estimation.}

In Fig.~\ref{fig:eff} (left), we report efficiency in terms of estimated exaFLOPs (EFLOPs). To ensure comparability, we estimate total training and sampling compute for each method based on their reported FLOPs per forward pass. Specifically, we use the following assumptions:
\begin{itemize}
    \item The FLOPs of a forward pass are reported by prior works (e.g., EDM\cite{karras2022elucidating}: $100$ GFLOPs, EDM2-S\cite{karras2024analyzing}: $102$ GFLOPs, SiT-XL\cite{ma2024sit}: $118.64$ GFLOPs).
    \item The FLOPs of a backward pass are measured empirically and are approximately $2\times$ the cost of a forward pass. (One JVP operation is also counted as a backward pass), say for our example on the first stage where we will perform one forward of the model and one JVP operation and one back propagation with JVP counted as one back propagation we have total flop amount of $1+ (2\times 2) = 1+4 $ forward flops.
    \item For training, the total compute is computed as:
    \begin{align*}
        \text{Total Train FLOPs} = &(\#\text{iters}) \times (\text{batch size}) \times (\text{forward} + \text{backward}) 
        \\
        &\times (\text{GFLOPs per fwd}).
    \end{align*}
    \item For sampling in the reflow process, the total compute is computed as:
    \begin{align*}
        \text{Total Sample FLOPs} = &(\#\text{samples}) \times (\#\text{steps}) \times (\text{forward passes per step}) 
        \\ &\times (\text{GFLOPs per fwd}).
    \end{align*}
    \item 
    \emph{Example: \Approach (Ours) on ImageNet-$64^2$.}
    For sampling, we require:
    \begin{align*}
        \underbrace{5 \times 10^6}_{\text{\#samples}} 
        \times \underbrace{63}_{\text{steps}} 
        \times \underbrace{2}_{\text{fwd/step (auto-guidance)}} 
        \times \underbrace{102}_{\text{GFLOPs/fwd}} \approx\;\; 64 \;\text{Eflops}.
    \end{align*}
    For training, we have two stages, with the first stage trained on the original flow and the second stage trained on the CFG velocity field:
    \begin{align*}
        &\underbrace{50{,}000}_{\text{iters}} 
        \times \underbrace{128}_{\text{batch}} 
        \times \underbrace{(1+4)}_{\text{fwd+back}} 
        \times \underbrace{102}_{\text{GFLOPs/fwd}} 
        \;\;
        \\ &+\;
        \underbrace{50{,}000}_{\text{iters}} 
        \times \underbrace{128}_{\text{batch}} 
        \times \underbrace{(3+4)}_{\text{fwd+back}} 
        \times \underbrace{102}_{\text{GFLOPs/fwd}} 
        \approx 8 \;\text{Eflops}.
    \end{align*}
    \item 
    \emph{Example: AYF \cite{sabour2025align}.}
    AYF does not require sampling, so only the training computation is considered:
    \begin{align*}
        &\underbrace{50{,}000}_{\text{iters}} 
        \times \underbrace{2048}_{\text{batch}} 
        \times \underbrace{(4+4)}_{\text{fwd+back}} 
        \times \underbrace{102}_{\text{GFLOPs/fwd}} 
        \approx 8.36 \times 10^{10} \;\; \text{GFLOPs} 
        \;\;\approx\;\; 84 \;\text{Eflops}.
    \end{align*}

\end{itemize}

\subsubsection{GPU Hours Estimation.}

In Fig.\ref{fig:eff}, we also estimate the total GPU hours for AYF\cite{sabour2025align} and 2-rectified flow++~\cite{lee2024improving}. For all methods, including ours, we follow the standard convention of computing GPU hours as
\[
\text{GPU Hours} = (\text{\# of GPUs}) \times (\text{wall-clock training time}).
\]
For example, \Approach{} requires 66 hours of wall-clock time on 8 A100 GPUs, yielding $8 \times 66 = 528$ GPU hours.

\begin{figure}[t]
    \includegraphics[width=1.\linewidth]{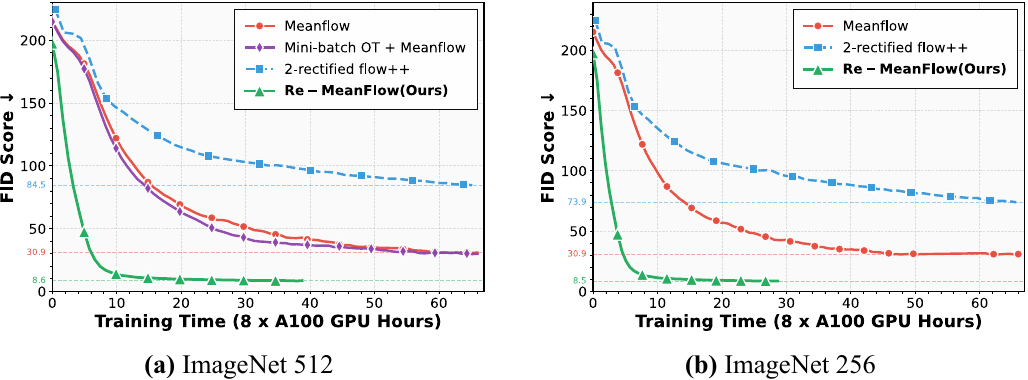}
    \caption{\textbf{Additional convergence comparison.}
        FID vs.\ training time (8\(\times\)A100 GPU hours) for \textbf{(a)} ImageNet \(512^2\) and \textbf{(b)} ImageNet \(256^2\).
        We compare \Approach{} against MeanFlow~\cite{geng2025mean} and closely related baselines, including 2-rectified flow++~\cite{lee2024improving} and MeanFlow trained with mini-batch OT couplings~\cite{tong2023improving} (512\(^2\) only).
        Across resolutions and baselines, \textbf{\Approach{} converges markedly faster and reaches substantially lower FID.}
    }
    \label{fig: eff2}
\end{figure}

\subsection{Additional Convergence Comparison with Other Related Methods}
\label{sec:convergence_ap}
In Fig.~\ref{fig: main}c, we compare the convergence of \Approach{} with MeanFlow~\cite{geng2025mean}. Here, we provide additional convergence results against closely related baselines.

\subsubsection{2-rectified flow++.}
We compare against 2-rectified flow++~\cite{lee2024improving}, which improves Rectified Flow but still models the \emph{instantaneous} velocity on rectified couplings, in contrast to \Approach{}, which models \emph{mean}-velocity. We observe that one-step sample quality under 2-rectified flow++ improves slowly. We attribute this to the sensitivity of one-step Euler updates to residual curvature: accurate one-step sampling with instantaneous velocity effectively requires near-perfect straightness, since even mild curvature can cause overshooting. Because rectified trajectories are not perfectly linear in practice, this sensitivity compounds and slows convergence. This observation aligns with our efficiency comparison in Sec.~\ref{sec: eff}: relative to 2-rectified flow++, \textbf{\Approach{} achieves a $33.4\%$ lower FID while being $26\times$ faster.}

\subsubsection{MeanFlow with Mini-batch OT couplings.}
We also train MeanFlow using Mini-batch OT\cite{tong2023improving} couplings, obtained by locally solving an OT matching within each mini-batch (via an OT server) to pair sampled noises and data points. Results on ImageNet $512^2$ are shown in Fig.~\ref{fig: eff2}a. While mini-batch OT slightly improves over standard MeanFlow, its convergence behavior remains similar and still lags substantially behind \Approach{}. This is consistent with our discussion in Sec.~\ref{sec: ot}: Mini-batch OT provides only local transport structure and does not guarantee globally straight trajectories.

Overall, \Approach{} converges substantially faster than these closely related alternatives, highlighting the synergy of modeling mean-velocity on rectified trajectories.

\begin{figure*}[t]
    \centering
    \includegraphics[width=1.\linewidth]{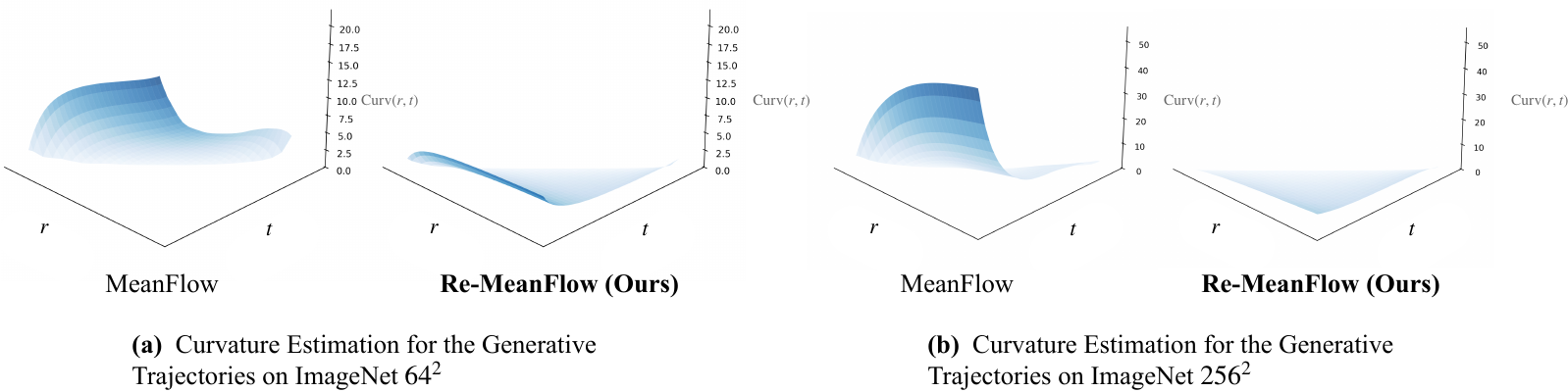}
    \caption{\textbf{Curvature estimation of generative trajectories} on ImageNet $64^2$ \textbf{(a)} and $256^2$ \textbf{(b)}. Using the proxy in Sec.~\ref{sec:curvature}, we visualize $\mathrm{Curv}(r,t)$ over the $(r,t)$-plane. \textbf{Rectified couplings used by \Approach{} induce substantially straighter trajectories (lower curvature) than the independent couplings used by MeanFlow.}}
    \label{fig: curv}
\end{figure*}

\subsection{Curvature Estimation}
\label{sec: curv_ab}
In Fig.~\ref{fig:method}, we visualize the curvature proxy on ImageNet $512^2$. Here we provide the corresponding curvature estimates at additional resolutions in Fig.~\ref{fig: curv}. The same trend holds across settings: compared to independent couplings, \textbf{the rectified couplings used to train \Approach{} consistently induce substantially lower curvature over the $(r,t)$-plane.}

\begin{figure*}[t]
    \centering
    \includegraphics[width=1.\linewidth]{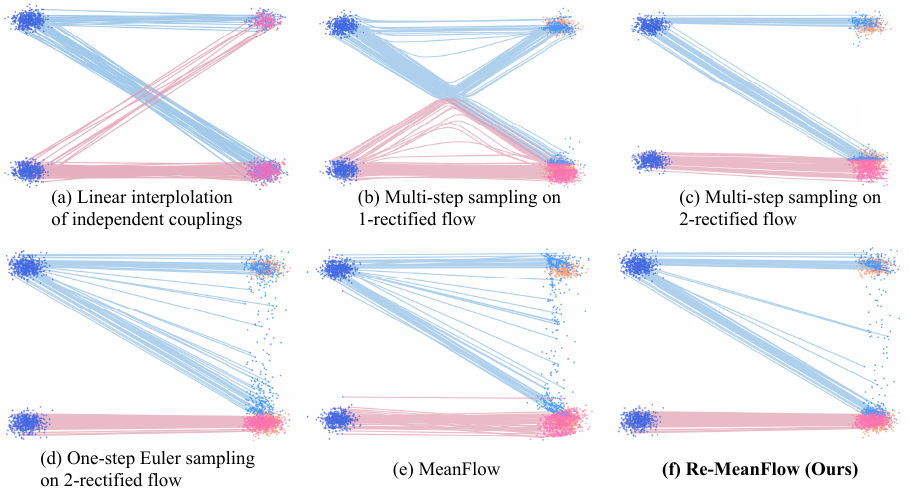}
    \caption{\textbf{A 2D Toy Example.} We consider a controlled 2D setup where a flow model transports a two-component Gaussian mixture on the left to a mixture on the right.
    \textbf{Panels (a-c):} \emph{(a)} Linear interpolation of independently sampled couplings 
    $p_{\mathbf x} \times p_{\mathbf z}$, which serves as the training signal 
    for the first velocity model.  
    \emph{(b)} The resulting 1-rectified flow learned from these independent 
    couplings; the learned velocity field remains noticeably curved.  
    \emph{(c)} Using the velocity field from (b), we generate a new set of 
    couplings and train a second velocity model on their linear interpolations, 
    yielding the 2-rectified flow.
    \textbf{Panel (d):} Due to imperfect straightening, one-step Euler sampling on the 2-rectified flow still yields noticeable outliers.
    \textbf{Panel (e):} MeanFlow trained directly on independent couplings fails to converge within the training budget because high-variance conditional velocities destabilize learning.
    \textbf{Panel (f): \Approach{} combines trajectory rectification with mean-velocity modeling, eliminating most outliers and achieving more accurate one-step generation.}}
    \label{fig:toy}
\end{figure*}

\begin{figure*}[t]
    \centering
    \includegraphics[width=1.\linewidth]{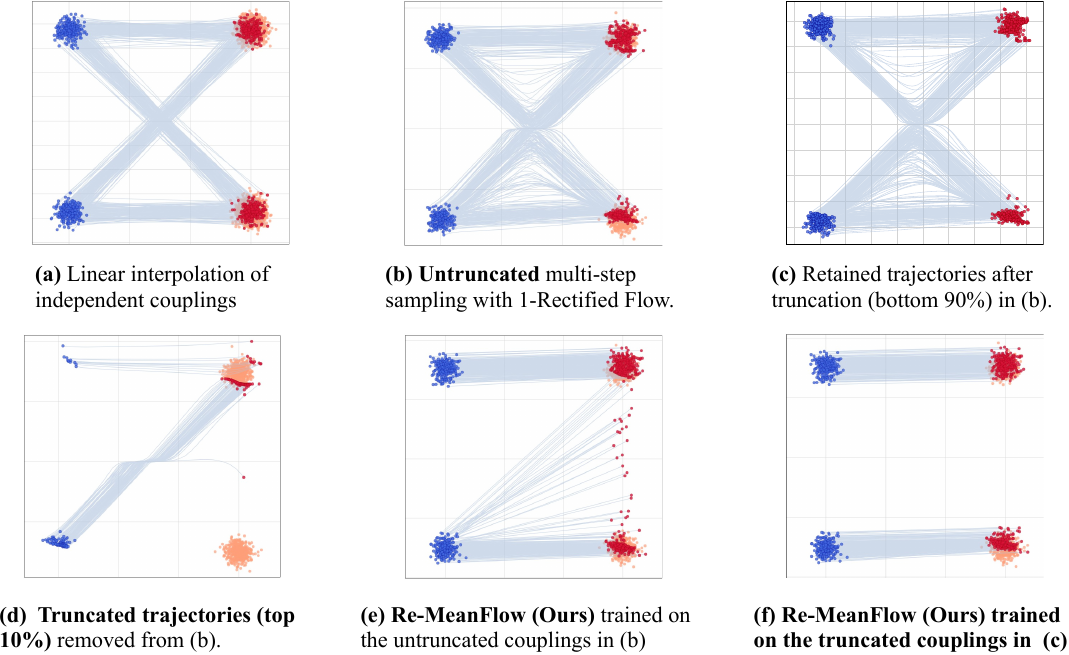}
    \caption{\textbf{Distance-based truncation removes high-curvature couplings and improves one-step behavior in a 2D toy task.}
    \textbf{(a)} Independent couplings visualized by linear interpolation between sampled endpoints.
    \textbf{(b)} Rectified couplings obtained by multi-step sampling with 1-rectified flow, where a small set of long-distance ``diagonal'' pairings induces high-curvature trajectories.
    \textbf{(c)} Retained trajectories after truncating the top \(10\%\) couplings ranked by endpoint distance \(\|\bx-\bz\|_2\) (bottom \(90\%\)).
    \textbf{(d)} Removed trajectories (top \(10\%\)) discarded by truncation.
    \textbf{(e)} \Approach{} trained on the untruncated rectified couplings in (b), exhibiting residual outliers.
    \textbf{(f) \Approach{} trained on the truncated couplings in (c), yielding much cleaner one-step transport with outliers largely eliminated.}}
    \label{fig:toy2}
\end{figure*}

\section{2D Toy Example}
To visualize the synergy of \Approach{} relative to MeanFlow\cite{geng2025mean} and Rectified Flow \cite{liu2022flow}, we present a 2D toy transport problem in Fig.~\ref{fig:toy}. The task maps a two-mode Gaussian mixture to another two-mode mixture. We compare one-step generation under a fixed budget of 20k training iterations for three approaches:
(i) 2-rectified flow,
(ii) MeanFlow, and
(iii) \Approach{} (ours), which trains a velocity model for 10k iterations to obtain a 1-rectified coupling, followed by 10k iterations of MeanFlow on the resulting couplings.

\paragraph{Fig.~\ref{fig:toy}d: 2-rectified flow.}
Even after one rectification step, some samples traverse \emph{disproportionately long distances} toward the denser mode (e.g., due to approximation error or mild mode imbalance). As a result, nearby noise samples can map to widely separated data points, yielding a large effective $L$ in Eq.~\ref{eq:lip} and inducing early-time curvature near $t\!\approx\!0$. The pronounced one-step deviations in Fig.~\ref{fig:toy}d illustrate this failure mode: one-step Euler updates using instantaneous velocity can overshoot when trajectories are not perfectly straight. Similar geometric imbalances can arise in real datasets, motivating robustness to residual curvature.

\paragraph{Fig.~\ref{fig:toy}e: MeanFlow.}
MeanFlow trained directly on independent couplings struggles to learn a coherent transport map within the same budget, resulting in poor one-step generation.

\paragraph{Fig.~\ref{fig:toy}f: \Approach{} (ours).}
\textbf{\Approach{} achieves more accurate one-step generation with substantially fewer invalid samples.} This illustrates the complementary roles of the two components: rectification reduces curvature enough to make MeanFlow optimization efficient, while mean-velocity modeling removes the requirement of perfectly straight trajectories.

\subsubsection{Distance-based Truncation}
To illustrate the effect of distance-based truncation, we perform an additional toy experiment shown in Fig.~\ref{fig:toy2}. We discard the top $10\%$ of couplings ranked by endpoint distance $\|\bx-\bz\|_2$. Fig.~\ref{fig:toy2}d visualizes the retained (truncated) trajectories. While this criterion does not guarantee removing \emph{all} high-curvature paths, it eliminates most of the ``diagonal'' couplings that exhibit the highest curvature in the toy setting (as analyzed above). Consequently,\textbf{ \Approach{} trained on the truncated couplings (Fig.~\ref{fig:toy2}f) achieves even cleaner one-step generation, removing nearly all outliers compared to training on the untruncated couplings (Fig.~\ref{fig:toy2}e).}

\newpage
\section{More Qualitative Results}
In this section, we present additional selected qualitative samples generated by \Approach{} across all ImageNet resolutions.
\subsection{ImageNet-$64^2$}
\begin{figure}[h]
    \centering
    \includegraphics[width=0.8\linewidth]{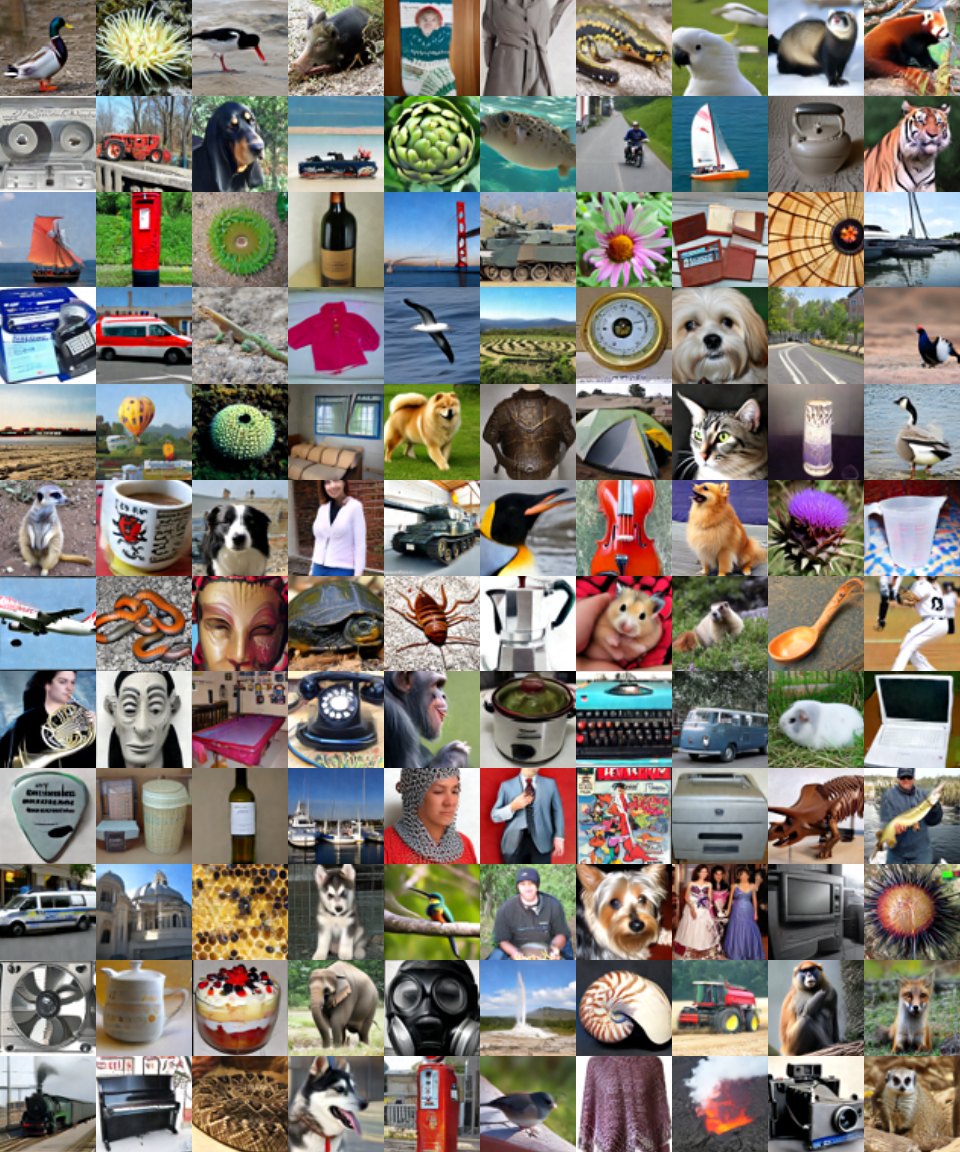}
    \caption{Selected qualitative results for \Approach (NFE=1) on ImageNet $64^2$.}
\end{figure}

\newpage
\subsection{ImageNet-$256^2$}
\begin{figure}[h]
    \centering
    \includegraphics[width=0.8\linewidth]{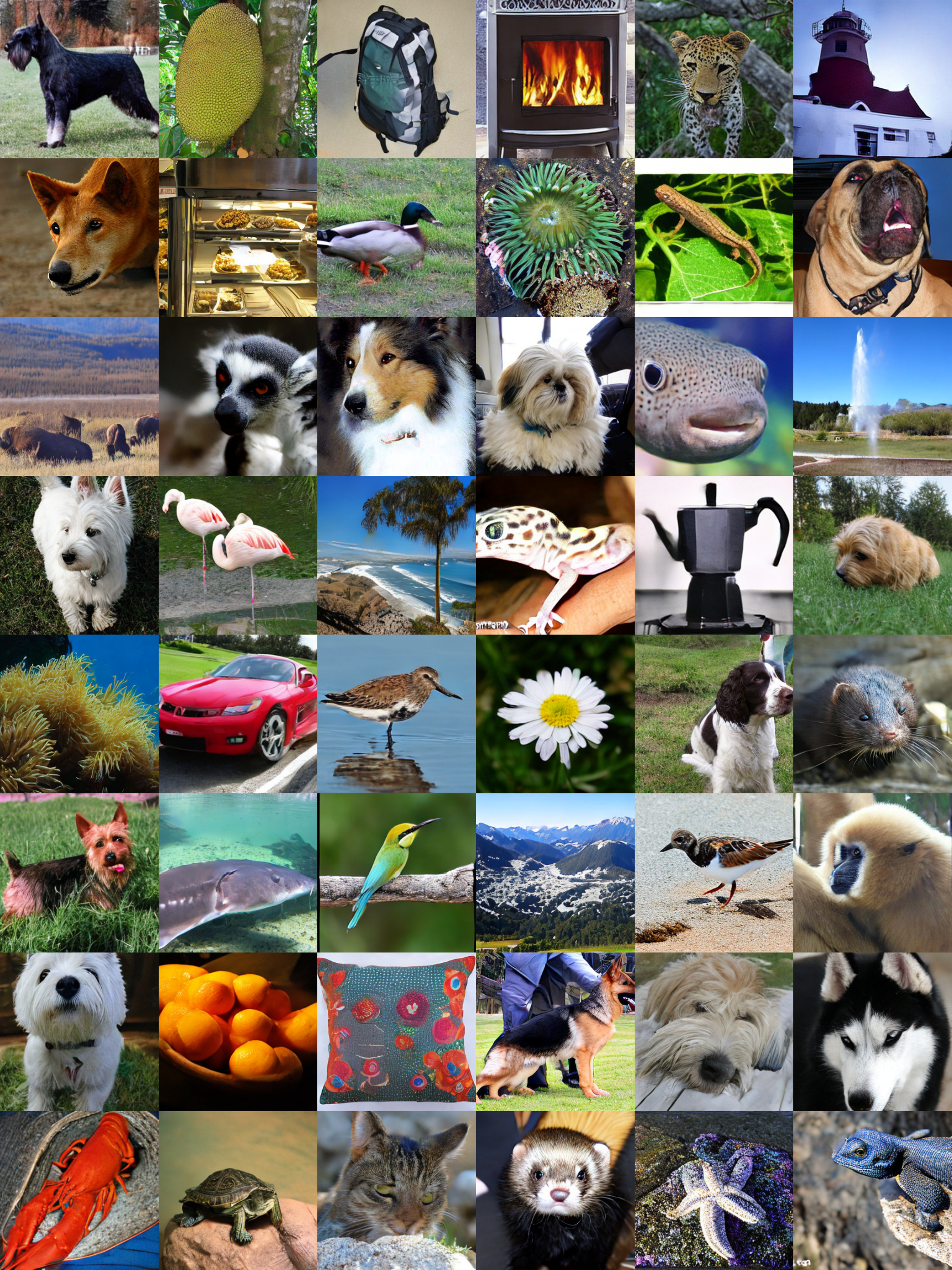}
    \caption{Selected qualitative results for \Approach (NFE=1) on ImageNet $256^2$.}
\end{figure}

\newpage
\subsection{ImageNet-$512^2$}
\begin{figure}[h]
    \centering
    \includegraphics[width=0.9\linewidth]{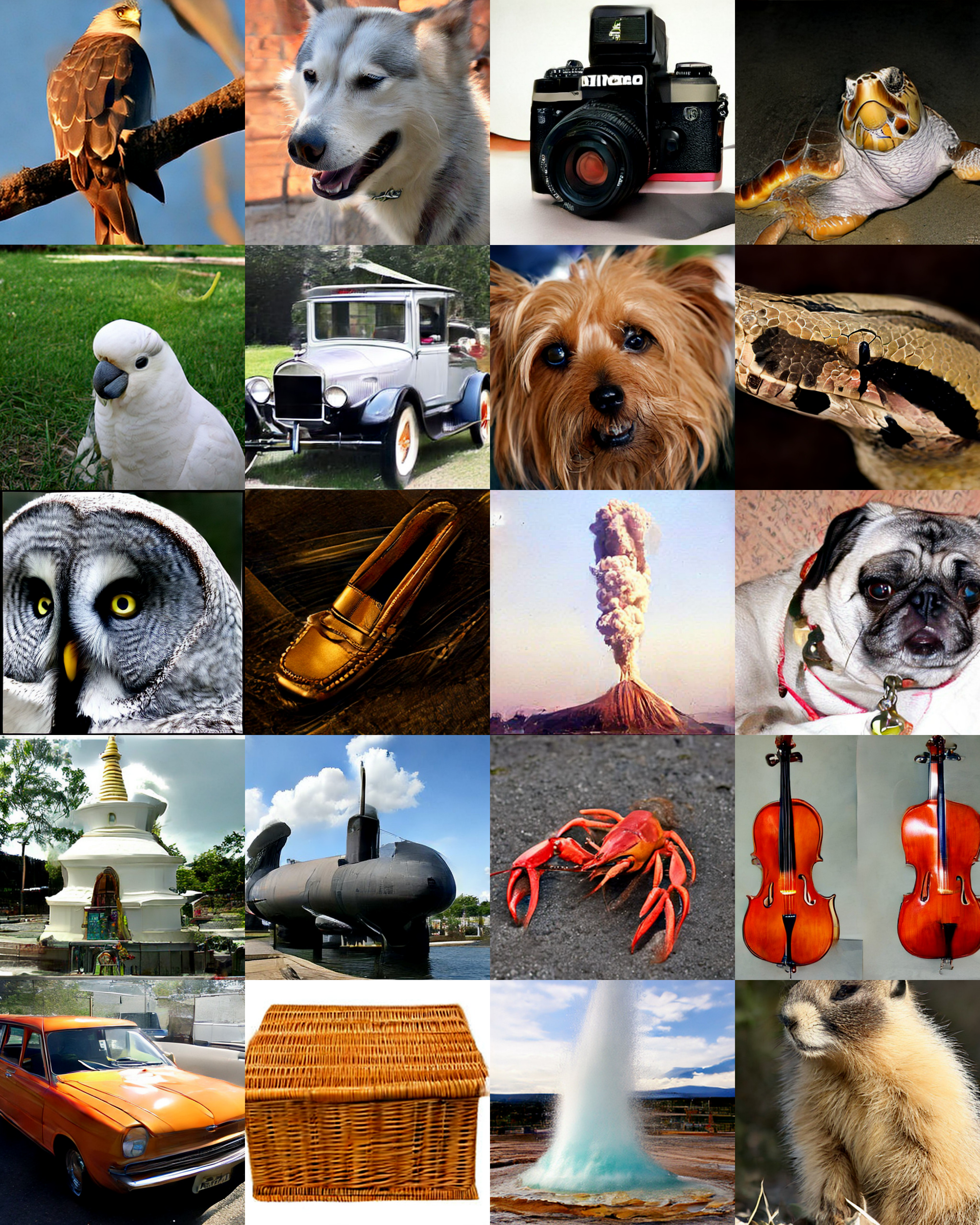}
    \caption{Selected qualitative results for \Approach (NFE=1) on ImageNet $512^2$.}
\end{figure}

%
%

\end{document}